\pdfoutput=1

\documentclass[11pt]{article}
\usepackage[hidelinks]{hyperref}

\usepackage[final]{acl}

\usepackage{times}
\usepackage{latexsym}

\usepackage[T1]{fontenc}

\usepackage[utf8]{inputenc}

\usepackage{microtype}

\usepackage{inconsolata}

\usepackage{graphicx}

\usepackage{amsmath}

\usepackage{booktabs}  
\usepackage{multirow}  
\usepackage{geometry}
\usepackage{makecell} 
\usepackage{natbib}
\usepackage{subcaption}
\usepackage{amssymb}
\usepackage{float}

\usepackage{stfloats}

\title{Bootstrapping Embeddings for Low Resource Languages}

\author{%
  Merve Basoz\textsuperscript{\,a}
  \and Andrew Horne\textsuperscript{\,b}
  \and Mattia Opper\textsuperscript{\,a} \\[0.5ex]
  \textsuperscript{a}School of Informatics, University of Edinburgh, 
   UK \\[0.5ex]
     \textsuperscript{b} Edina, University of Edinburgh,
   UK \\[0.5ex]
  \texttt{\{s2451985, ahorne2, m.opper\}@ed.ac.uk}
}

\begin{document}
\maketitle
\begin{abstract}
Embedding models are crucial to modern NLP. However, the creation of the most effective models relies on carefully constructed supervised finetuning data. For high resource languages, such as English, such datasets are readily available. However, for hundreds of other languages, they are simply non-existent. We investigate whether the advent of large language models can help to bridge this gap. We test three different strategies for generating synthetic triplet data used to optimise embedding models. These include in-context learning as well as two novel approaches, leveraging adapter composition and cross lingual finetuning of the LLM generator (XL-LoRA) respectively. We find that while in-context learning still falls short of strong non-synthetic baselines, adapter composition and XL-LoRA yield strong performance gains across a wide array of tasks and languages, offering a clear, scalable pathway to producing performant embedding models for a wide variety of languages.
\end{abstract}

\section{Introduction}
In natural language processing (NLP), embeddings play an important role, powering crucial applications such as retrieval augmented generation, semantic matching and classification, among others \citep{cer2018universalsentenceencoder, rag, thakur2021beirheterogenousbenchmarkzeroshot, muennighoff2022mteb}. Creating the most effective embedding models relies on human annotated triplet data, which allows the model to learn complex semantic relationships during finetuning \citep{sbert, ni-etal-2022-sentence}. However, such datasets are costly to produce and although they are abundant in high resource languages such as English, they are simply unavailable for many languages, despite these languages often having many millions of speakers and high demand for NLP technologies. While unsupervised approaches have shown promise as a potential solution, their performance still lags substantially behind what can be achieved using curated supervision \citep{simcse, wu-etal-2022-pcl}. 

The rise of large language models holds the promise of bridging this gap. Given their impressive broad spectrum capabilities \cite{bubeck2023sparksartificialgeneralintelligence}, can we use them to generate high-quality synthetic data, cheaply and effectively bridging the gap for languages and domains where human annotated resources simply do not exist? Recent work, investigating synthetic data generation \citep{syncse}, has shown promising results on standard English benchmarks. Here we examine whether it can be applied to low resource languages where the impact is arguably far greater, as alternatives often simply do not exist. 

We examine the efficacy of three approaches to synthetic data generation: one uses in-context learning following prior work \cite{syncse} and two novel methods; one in which we optimise the LLM data generator using adapter composition and the other in which we use a specialised cross lingual adaptation regime we dub XL-LoRA\footnote{Code available at: \mbox{\url{github.com/mbasoz/xllora-embedding}}}.  We find that these latter two approaches show clear consistent gains against strong non-synthetic baselines across a wide array of tasks and languages. Moreover, XL-LoRA, our best performing method, requires no data in the target language to optimise the generator and produces highly competitive results from just a small scale finetuning dataset, though it can in principle be scaled further. We believe this offers promising pathway for producing performant models for the myriad languages where the requisite resources would otherwise be unavailable.



\section{Background and Related Work}



\noindent{\textbf{Transformer Embeddings:}} Early encoder-only transformers such as BERT \cite{devlin-etal-2019-bert} produced poor embedding models, underperforming simple word embeddings despite being much more powerful in theory \cite{sbert}. This was in large part due to a tendency to produce anisotropic embedding spaces \cite{bert-whitening, trafo-anisotropy}, which is provably harmful for learning useful representations \cite{ua}. However, these limitations were short-lived. Seminal work by \citet{sbert} demonstrated that by harnessing supervised NLI datasets and an appropriate representation level objective, transformer embeddings could achieve a new state of the art. This discovery was quickly followed by the SimCSE framework of \citet{simcse}. In SimCSE representations are optimised using a contrastive objective inspired by advances in computer vision \cite{chen2020simpleframeworkcontrastivelearning}. The objective requires maximising the cosine similarity between an anchor sentence and a positive example, while simultaneously minimising similarity to a set of negative examples. It is defined as follows:

\begin{equation}
- \log \frac{e^{\text{sim}(h_i, h_i^+)/\tau}}{\sum_{j=1}^{M} \left( e^{\text{sim}(h_i, h_j^+)/\tau} + e^{\text{sim}(h_i, h_j^-)/\tau} \right)}.
\end{equation}

\noindent Where $h_{i}$ denotes the anchor, $h_{i}^{+}$ denotes the positive target and $h_{i}^{-}$ denotes the hard negative. The objective can be applied in an unsupervised manner, in which case the term including $h_{i}^{-}$ is omitted, and the positive simply corresponds to the anchor with a different dropout mask applied. However, performance improves significantly when the objective is applied in a supervised setting, where for each example sentence a human annotated positive and hard negative is provided forming a triplet. The SimCSE framework set a new state of the art and has remained dominant since. However, its success is crucially dependent on the existence of high-quality triplet data, with the best reported performance coming from triplets obtained via the NLI dataset \cite{bowman-etal-2015-large}.\\
\newline
\noindent\textbf{Embeddings for Low Resource Languages:}
A core challenge of the transformer-based approach is its reliance on supervision and scale, making it difficult to apply to low resource languages.
One approach to this issue is to use lightweight models that can be trained fully unsupervised \cite{mao2023leallalearninglightweightlanguageagnostic, bestgen-2024-satlab, straesemeval, banyan}. However, these models have weaker capacity than large scale transformers, which places a ceiling on their efficacy. 
A second approach is to attempt to train multilingual models that have aligned embedding spaces, so that knowledge of a high resource language can at least partially transfer to a low resource one. This can be achieved implicitly through the inclusion of multiple languages in the pre-training data and careful control over sampling from each \cite{conneau-etal-2018-xnli, mmbert}. Or explicitly through the use of large-scale parallel corpora \cite{muse, labse, parascale, laser3}. 
Aligned embedding spaces enable cross lingual transfer \cite{reimers-gurevych-2020-making, colbert-x}, whereby such models can be finetuned on high quality supervised data in one language and then partially transfer the same capability to the target language. This approach can produce very strong results. However, it remains suboptimal compared with the ideal situation where high-quality supervised data is available in the target language. \\ 

\noindent\textbf{Data Synthesis Using LLMs:}
As the capabilities of LLMs grow, they have begun to gain ever-increasing traction as tools for data synthesis. Arising from the hope that the sheer amount of knowledge they have internalised can be used to address the challenge of obtaining high-quality data for the multitudes of tasks and domains that cover NLP \citep{hartvigsen-etal-2022-toxigen, sahu-etal-2022-data, wang-etal-2023-self-instruct, honovich-etal-2023-unnatural}. Most relevant for our work, \citet{syncse} introduce SynCSE: a technique for synthetically generating triplet data required to train effective embedding models. To do so they use five-shot in-context learning, with alternating prompts and examples and either generate the positive and negative examples conditioned on a provided anchor sentence or generate all three components of the triplet simultaneously. \citet{syncse} demonstrated strong results on standard English benchmarks, leaving open the promise that the same technique could be applied to generate data for specialised domains or language adaptation. \\

\noindent\textbf{Summary:}  When coupled with high-quality triplet data, transformer encoders can produce powerful, effective embeddings. For high resource languages like English, where such datasets abound, there exist a vast array of effective embedding models. However, for low resource languages, where such data is lacking, producing embeddings remains a challenge. We explore whether the advent of powerful decoder only language models can allow us to bridge this gap. Can we synthesise high-quality examples for low resource languages, and what are the best approaches to do so?

\section{Pipeline and Methodology}
\label{sec:data_generation}

\begin{figure}[tbp]
\centering
\includegraphics[width=0.45\textwidth]{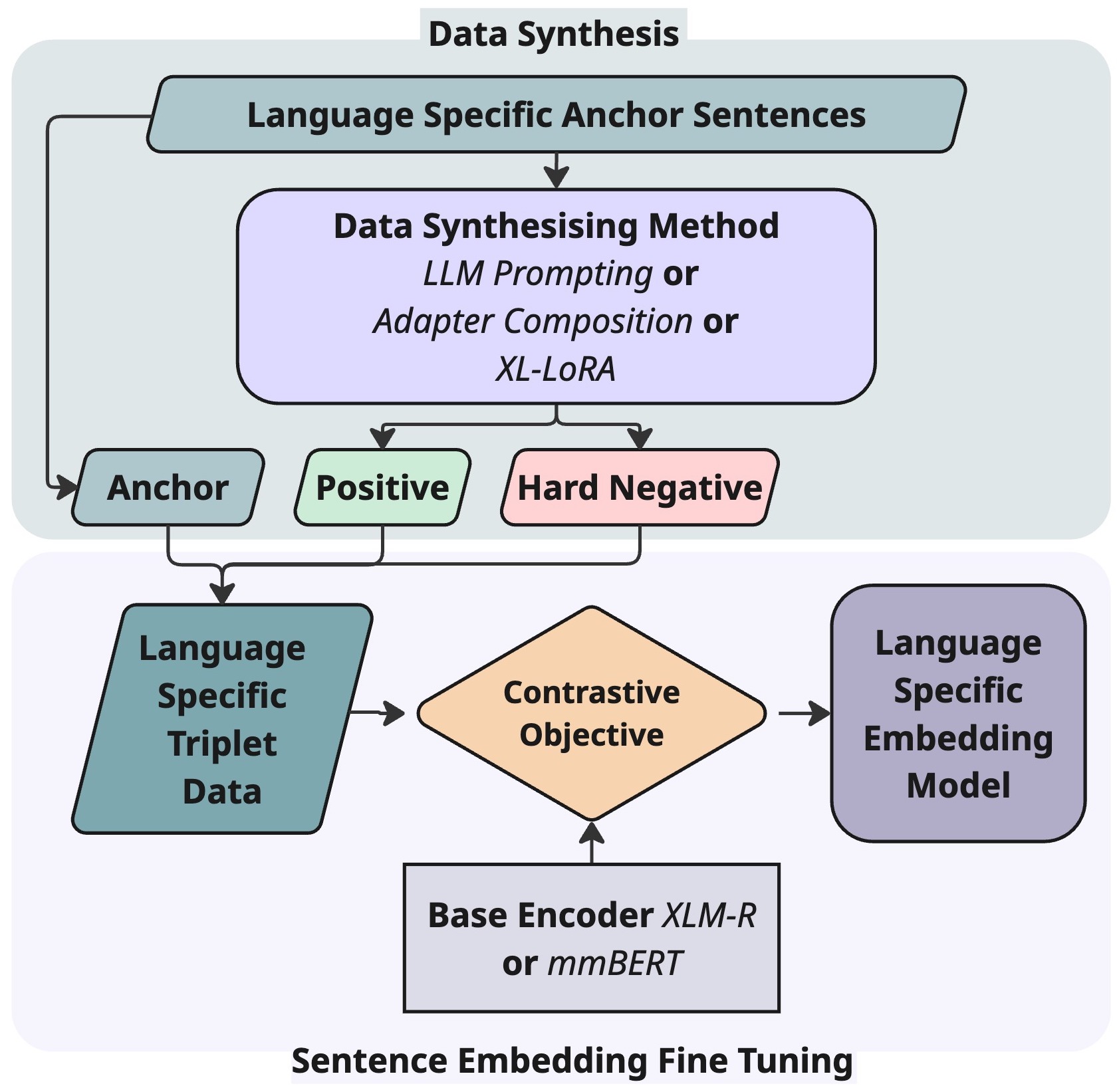}
\caption{Pipeline overview: data is synthesised using an LLM, the result of which is then used to finetune an encoder, resulting in the final embedding model.}
\label{fig:sentence_embedding_ft}
\end{figure}

In this section, we outline a) the overall data generation pipeline and subsequent training of the embedding model and b) the different techniques used to optimise the LLM data generator. 

\subsection{Pipeline}
Our data synthesis pipeline consists of three core stages. First, we collect anchor sentences for the target language from widely available web corpora. Second, we pass these sentences to the LLM generator which outputs the corresponding positive and hard negative pairs for the anchor, forming the triplet. Finally, once we have created the entire dataset, we then use it to optimise our transformer encoder backbone using the supervised SimCSE contrastive objective \cite{simcse}. The resulting embedding model is then evaluated zeroshot on downstream tasks. Figure \ref{fig:sentence_embedding_ft} shows an overview of the pipeline. For all experiments, we used Gemma 3 27b \cite{gemma_2025} as the generator, which we found to be the most effective multilingual model given our compute budget. \\ 


\noindent\textbf{In-Context Learning with Prompting:} Our first approach follows that of SynCSE \cite{syncse}, which demonstrated that with fewshot prompting applied to English, language models can generate effective synthetic data for training embedding models, outperforming unsupervised baselines and approaching parity with human annotations. Motivated by these findings, we tested whether SynCSE can be applied to low resource languages, where  the need for synthetic data is arguably far higher. An overview of the exact pipeline for generating prompts can be found in Appendix \ref{subsec:localised-syncse-partial} while the prompts themselves can be found in \ref{subsec:SynCSE-prompts}. Our approach closely mirrors the original SynCSE, though it differs in two regards. First, we used Gemma 3 27b rather than GPT 3.5 for parity with our other experiments. Second, we instruct the model to provide the output in the target language while the examples are in English. We also experimented with translating the prompt and examples, but found this to have a negligible effect compared to using English. \\


\noindent\textbf{Adapter Composition:} Beyond simple prompting, we also explored whether optimising the LLM generator itself could prove beneficial. After all, producing quality triplets requires an excellent grasp of semantics, and on top of that, the model must also be able to apply this knowledge to a low resource language. To this end, we turned to LoRA \cite{hu2021loralowrankadaptationlarge} finetuning, as it has demonstrated high efficacy in low-data training regimes \cite{whitehouse2024lowrankadaptationmultilingualsummarization} and provides minimal computational overhead. A further desirable characteristic of LoRA is that adapters can be composed, allowing for flexible combination of different capabilities \citep{zhang2023composingparameterefficientmodulesarithmetic, zhao2024adamergexcrosslingualtransferlarge}. This is particularly desirable in our case because we are looking to optimise for two characteristics: a) the model must be able to understand what constitutes a useful positive or negative example for a given anchor, and b) it must have competency in the target language to generate grammatical outputs. As a result, we turn to AdamergeX. A technique recently introduced by \citet{zhao2024adamergexcrosslingualtransferlarge}, which showed strong results in cross lingual transfer in key tasks, including natural language understanding, summarisation, and reasoning. The core equation underlying AdamergeX composition is given as follows:

\begin{equation}
\resizebox{0.9\columnwidth}{!}{$\displaystyle
\underbrace{\mathbf{TA}_{tgt}}_{\text{target task}} = \underbrace{\mathbf{TA}_{src}}_{\text{source triplet tuning}} \overbrace{+}^{\text{elem-wise}} \underbrace{\lambda \left( \mathbf{LA}_{tgt} - \mathbf{LA}_{src} \right)}_{\text{Generic Instruction Tuning}}
$}
\end{equation}


Shown above, AdamergeX requires training three separate adapters: an adapter for the task in the source language and two generic language adapters, one in the source language and one in the target. The source language adapter is subtracted from that of the target to remove any artifacts outside of pure linguistic competency and is then added to the task adapter, with a hyperparameter lambda controlling the impact of the language adapter on the composition. In our case, the task is triplet generation and we train the corresponding adapter using a subset of English NLI \cite{williams-etal-2018-broad, bowman-etal-2015-large}. For training the task adapter we use an instruction tuning format, inputting premise and outputting contradiction or entailment sentences. The prompts to convert the task to the requisite format can be found in Appendix~\ref{subsec:prompting_adamergex}. Language adapters are trained using causal language modeling on the Aya dataset. We used either the core dataset or the machine translated collections split based on availability \citep{singh2024aya}. Following ablations, which can be found in \ref{subsec:adamergex-ablations}, we observed that training two separate Task Adapters, for generating the positive and hard negative respectively, improved performance compared with using a single adapter for both. Further ablations over hyperparameters can be found in Appendix~\ref{subsec:prompting_adamergex}. Finally, following \citet{zhao2024adamergexcrosslingualtransferlarge} and due to the limited size of Aya, we used 10k examples to train each adapter. \\

\begin{figure}[t]
\centering
\includegraphics[width=0.5\textwidth]{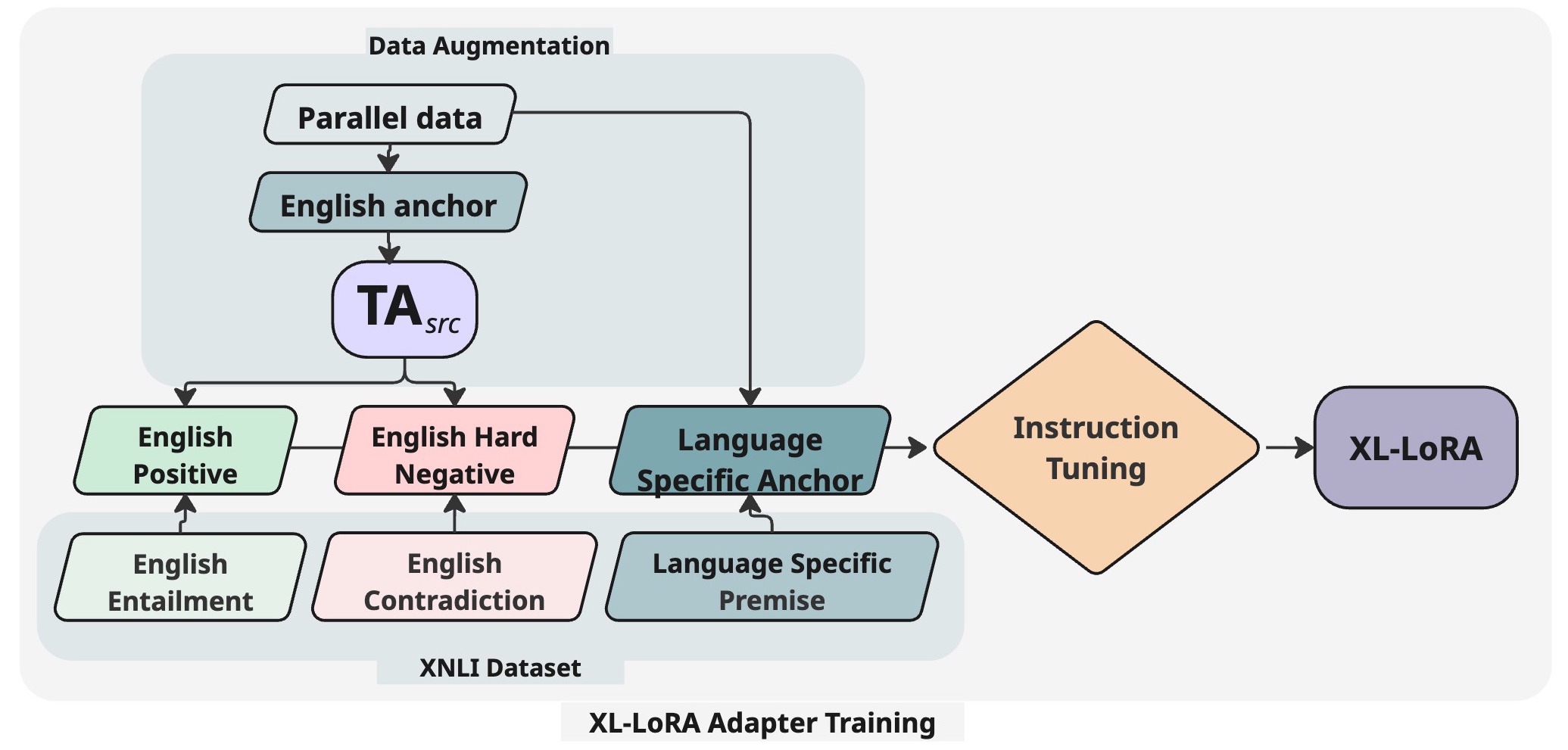}
\caption{XL-LoRA training data construction. Starting with high quality translations we generate positives and negatives based on English, before swapping back the anchor to the original non-English language. Resulting examples are used to finetune the XL-LoRA generator.}
\label{fig:clloramix}
\end{figure}

\noindent\textbf{XL-LoRA:} For our final approach, we introduce a technique we term XL-LoRA (cross lingual LoRA). Here we do not require the generator to produce outputs in the target language. Rather we generate English positives and negatives while only the anchor remains in the target. This approach is motivated by two key observations. First, LLMs display internal cross lingual alignment \cite{wendler-etal-2024-llamas}. Whereby in the middle layers tokens frequently occupy an abstract concept space, shared across languages. Second, LLMs have a strong bias towards outputting English. This means that even if they understand a task completely, the very fact of having to output non-English tokens can be highly detrimental to performance, masking true capabilities and perhaps partially being responsible for the capability gap observed in low resource languages \cite{pomerenke2025ailanguageproficiencymonitor, 
ahuja2023megamultilingualevaluationgenerative}.  Consequently, we constructed an additional finetuning set where anchors are in the target language, but the positives and negatives (i.e. generation targets for the LLM) are in English. We used the following steps to create this dataset as illustrated in Figure \ref{fig:clloramix}. \\

\noindent\textbf{Step 1: } First, we sourced high-quality translations between English and non-English languages\footnote{\scriptsize
Training of the XL-LoRA adapter includes the following languages: Arabic, Bulgarian, German, Greek, Spanish, French, Hindi, Russian, Swahili, Thai, Turkish, Urdu, Vietnamese and Chinese.}. One of these is the XNLI test set \cite{conneau-etal-2018-xnli}, which unlike the XNLI training set, contains translations by human experts. In this case, we already had positives and negatives available for each example so we simply swapped the English anchor for multilingual ones. As a result, the languages used to train the adapter are restricted to those present in XNLI and are largely differ from the target languages used as the anchor sentences later, with Hindi as the sole exception. This avoids the need to include target language supervision, which is often difficult to obtain for low-resource languages. \\

\noindent\textbf{Step 2: } To further increase the number of examples to 10k, matching the  SFT set sizes used for AdaMergeX, we could not rely on the XNLI test set alone, as it is very limited in scale. For further data, we found high-quality human translation data \cite{tiedemann-de-gibert-2023-opus} in the same languages covered by XNLI. We then used our triplet generation model, finetuned on English, to generate the corresponding positives and negatives before swapping out the English anchor with its multilingual equivalent. We found that high-quality translations in the XL-LoRA training were absolutely crucial to ensuring success (ablations can be found later in the paper), and using machine translated examples was highly detrimental. However, we note that this is only required for the small scale finetuning of the generator. Additionally, we emphasize that, unlike in the adapter composition approach, no data in the target language is required for training the XL-LoRA adapter, and the vast majority of the languages we later evaluate on are not present in the training data. \\

\noindent\textbf{Step 3: } We train the XL-LoRA adapter using instruction tuning, where the model is prompted in a zero shot manner using the positive and negative pair generation instructions described in Appendix~\ref{subsec:prompting_adamergex}. Each prompt takes a non-English anchor sentence as a premise and instructs the model to generate either a contradictory sentence as a hard negative or an entailment sentence as a positive in English. During training, the reference positives and hard negatives are provided only as target outputs for supervision and are not included in the input prompt. Once this is completed, the generator can then be deployed at scale for data synthesis. \\

\noindent\textbf{Summary:} We present three methods for optimising our synthetic data generator: one based on prompting, the second on adapter composition and the final uses cross lingual examples. Next, we test them experimentally. 

\section{Experiments}
\noindent\textbf{Setup:} We want to evaluate the efficacy of synthetic data for training embedding models suitable for low resource languages. To do so we select two popular and capable multilingual models, XLM-R \citep{conneau2020unsupervisedcrosslingualrepresentationlearning} and mmBERT \citep{mmbert}, as backbones that we finetune into embedding models.
We compare against the following baselines not reliant on synthetic data:
\begin{enumerate}
  \item \textbf{Base Encoder:} The first baselines are simply the pretrained backbones without any finetuning. These are unlikely to be good embedding models out of the box \cite{sbert}, but provide a useful lower bound.
  \item \textbf{Unsupervised:} Here we finetune backbones using the unsupervised SimCSE objective \cite{simcse} on unlabeled data in the target language. This approach mitigates the lack of human annotated data, and has been shown to produce strong results on English STS.
  \item \textbf{Cross Lingual:} While unsupervised SimCSE is a strong method for training embedding models its performance still lags in comparison to human annotated data. An alternative approach is to take advantage of the fact that multilingual encoders learn partially shared representations across languages \cite{conneau2020unsupervisedcrosslingualrepresentationlearning}, which enables a good degree of zeroshot cross lingual transfer. Therefore, a highly performant alternative approach is to finetune using human annotated English data, and then evaluate zeroshot transfer to the target language \cite{colbert-x}. 
\end{enumerate}

\noindent We compare these baselines against the three approaches for synthetic data generation described in Section~\ref{sec:data_generation}. Namely, prompting using in-context examples, adapter composition and finally XL-LoRA. The cross lingual baseline is trained using supervised SimCSE on the English NLI dataset from \citet{williams-etal-2018-broad, bowman-etal-2015-large}. The unsupervised baseline for each target language is trained using sentences sourced from the Leipzig Corpora Collection \citep{goldhahn-etal-2012-building} except for the Hausa data sourced from Opus \cite{nygaard2003opus}, while the synthetic approaches use these same sentences as anchors and then generate the corresponding hard positives and negatives - totalling 275k training examples, in line with the English NLI sample size. We evaluate performance using both STS/STR tasks from \cite{semreltask} and a subset of retrieval tasks from MTEB \cite{muennighoff2022mteb} specifically focusing on low resource languages. Further training and hyperparameter details can be found in Appendix~\ref{subsubsec:training_details}. 
\\

\begin{table*}[t]
\centering
\scriptsize
\setlength{\tabcolsep}{2pt}
\renewcommand{\arraystretch}{0.8}
\resizebox{\textwidth}{!}{%
\begin{tabular}{lcccccccc@{\hspace{6pt}\vrule\hspace{6pt}}cccccccc}
\toprule
 & \multicolumn{8}{c}{\textbf{XLM-R}} & \multicolumn{8}{c}{\textbf{MM-BERT}} \\
\cmidrule(lr){2-9}\cmidrule(lr){10-17}
\textbf{Method} & \textbf{Afr} & \textbf{Hin} & \textbf{Mar} & \textbf{Tel} & \textbf{Ind} & \textbf{Hau} & \textbf{Kor} & \textbf{Score}
               & \textbf{Afr} & \textbf{Hin} & \textbf{Mar} & \textbf{Tel} & \textbf{Ind} & \textbf{Hau} & \textbf{Kor} & \textbf{Score} \\
\midrule
Base Encoder  
& 56.2 & 52.7 & 55.7 & 46.3 & 46.7 & 4.1 & 60.8 & 46.1
& 72.0 & 63.8 & 70.7 & 66.8 & \textbf{52.2} & 21.7 & 62.0 & 58.5 \\
\midrule
Unsupervised  
& \begin{tabular}[c]{@{}c@{}}74.8\\{\scriptsize$\pm$0.3}\end{tabular}
& \begin{tabular}[c]{@{}c@{}}69.7\\{\scriptsize$\pm$0.2}\end{tabular}
& \begin{tabular}[c]{@{}c@{}}77.1\\{\scriptsize$\pm$0.2}\end{tabular}
& \begin{tabular}[c]{@{}c@{}}76.2\\{\scriptsize$\pm$0.3}\end{tabular}
& \begin{tabular}[c]{@{}c@{}}39.2\\{\scriptsize$\pm$0.3}\end{tabular}
& \begin{tabular}[c]{@{}c@{}}45.4\\{\scriptsize$\pm$0.7}\end{tabular}
& \begin{tabular}[c]{@{}c@{}}70.9\\{\scriptsize$\pm$0.3}\end{tabular}
& \begin{tabular}[c]{@{}c@{}}64.7\\{\scriptsize$\pm$0.2}\end{tabular}
& \begin{tabular}[c]{@{}c@{}}75.5\\{\scriptsize$\pm$1.2}\end{tabular}
& \begin{tabular}[c]{@{}c@{}}58.0\\{\scriptsize$\pm$2.6}\end{tabular}
& \begin{tabular}[c]{@{}c@{}}68.6\\{\scriptsize$\pm$1.3}\end{tabular}
& \begin{tabular}[c]{@{}c@{}}62.1\\{\scriptsize$\pm$3.5}\end{tabular}
& \begin{tabular}[c]{@{}c@{}}44.7\\{\scriptsize$\pm$1.7}\end{tabular}
& \begin{tabular}[c]{@{}c@{}}25.5\\{\scriptsize$\pm$1.2}\end{tabular}
& \begin{tabular}[c]{@{}c@{}}58.0\\{\scriptsize$\pm$2.9}\end{tabular}
& \begin{tabular}[c]{@{}c@{}}56.1\\{\scriptsize$\pm$0.8}\end{tabular} \\
\midrule
Cross Lingual  
& \begin{tabular}[c]{@{}c@{}}78.0\\{\scriptsize$\pm$0.3}\end{tabular}
& \begin{tabular}[c]{@{}c@{}}77.4\\{\scriptsize$\pm$0.1}\end{tabular}
& \begin{tabular}[c]{@{}c@{}}81.6\\{\scriptsize$\pm$0.2}\end{tabular}
& \begin{tabular}[c]{@{}c@{}}80.9\\{\scriptsize$\pm$0.7}\end{tabular}
& \begin{tabular}[c]{@{}c@{}}47.1\\{\scriptsize$\pm$0.6}\end{tabular}
& \begin{tabular}[c]{@{}c@{}}48.2\\{\scriptsize$\pm$1.0}\end{tabular}
& \begin{tabular}[c]{@{}c@{}}\textbf{79.9}\\{\scriptsize$\pm$0.1}\end{tabular}
& \begin{tabular}[c]{@{}c@{}}70.4\\{\scriptsize$\pm$0.2}\end{tabular}
& \begin{tabular}[c]{@{}c@{}}78.0\\{\scriptsize$\pm$0.4}\end{tabular}
& \begin{tabular}[c]{@{}c@{}}77.7\\{\scriptsize$\pm$0.1}\end{tabular}
& \begin{tabular}[c]{@{}c@{}}79.8\\{\scriptsize$\pm$0.4}\end{tabular}
& \begin{tabular}[c]{@{}c@{}}73.6\\{\scriptsize$\pm$0.6}\end{tabular}
& \begin{tabular}[c]{@{}c@{}}49.3\\{\scriptsize$\pm$0.6}\end{tabular}
& \begin{tabular}[c]{@{}c@{}}29.4\\{\scriptsize$\pm$1.7}\end{tabular}
& \begin{tabular}[c]{@{}c@{}}\textbf{79.6}\\{\scriptsize$\pm$0.1}\end{tabular}
& \begin{tabular}[c]{@{}c@{}}66.8\\{\scriptsize$\pm$0.4}\end{tabular} \\
\midrule
Synth - Prompting  
& \begin{tabular}[c]{@{}c@{}}\textbf{81.4}\\{\scriptsize$\pm$0.5}\end{tabular}
& \begin{tabular}[c]{@{}c@{}}77.9\\{\scriptsize$\pm$0.3}\end{tabular}
& \begin{tabular}[c]{@{}c@{}}82.3\\{\scriptsize$\pm$0.3}\end{tabular}
& \begin{tabular}[c]{@{}c@{}}81.4\\{\scriptsize$\pm$0.5}\end{tabular}
& \begin{tabular}[c]{@{}c@{}}38.3\\{\scriptsize$\pm$0.9}\end{tabular}
& \begin{tabular}[c]{@{}c@{}}47.4\\{\scriptsize$\pm$1.2}\end{tabular}
& \begin{tabular}[c]{@{}c@{}}69.9\\{\scriptsize$\pm$0.5}\end{tabular}
& \begin{tabular}[c]{@{}c@{}}68.4\\{\scriptsize$\pm$0.2}\end{tabular}
& \begin{tabular}[c]{@{}c@{}}\textbf{80.6}\\{\scriptsize$\pm$0.3}\end{tabular}
& \begin{tabular}[c]{@{}c@{}}76.8\\{\scriptsize$\pm$0.4}\end{tabular}
& \begin{tabular}[c]{@{}c@{}}76.3\\{\scriptsize$\pm$0.6}\end{tabular}
& \begin{tabular}[c]{@{}c@{}}76.6\\{\scriptsize$\pm$0.3}\end{tabular}
& \begin{tabular}[c]{@{}c@{}}41.3\\{\scriptsize$\pm$1.0}\end{tabular}
& \begin{tabular}[c]{@{}c@{}}45.2\\{\scriptsize$\pm$1.4}\end{tabular}
& \begin{tabular}[c]{@{}c@{}}68.0\\{\scriptsize$\pm$0.4}\end{tabular}
& \begin{tabular}[c]{@{}c@{}}66.4\\{\scriptsize$\pm$0.4}\end{tabular} \\
\midrule
Synth - Adapter Composition  
& \begin{tabular}[c]{@{}c@{}}80.4\\{\scriptsize$\pm$0.3}\end{tabular}
& \begin{tabular}[c]{@{}c@{}}77.6\\{\scriptsize$\pm$0.2}\end{tabular}
& \begin{tabular}[c]{@{}c@{}}\textbf{84.8}\\{\scriptsize$\pm$0.2}\end{tabular}
& \begin{tabular}[c]{@{}c@{}}83.0\\{\scriptsize$\pm$0.2}\end{tabular}
& \begin{tabular}[c]{@{}c@{}}46.4\\{\scriptsize$\pm$0.3}\end{tabular}
& \begin{tabular}[c]{@{}c@{}}49.5\\{\scriptsize$\pm$0.5}\end{tabular}
& \begin{tabular}[c]{@{}c@{}}74.7\\{\scriptsize$\pm$0.6}\end{tabular}
& \begin{tabular}[c]{@{}c@{}}70.9\\{\scriptsize$\pm$0.2}\end{tabular}
& \begin{tabular}[c]{@{}c@{}}79.9\\{\scriptsize$\pm$0.3}\end{tabular}
& \begin{tabular}[c]{@{}c@{}}77.6\\{\scriptsize$\pm$0.2}\end{tabular}
& \begin{tabular}[c]{@{}c@{}}82.2\\{\scriptsize$\pm$0.4}\end{tabular}
& \begin{tabular}[c]{@{}c@{}}81.1\\{\scriptsize$\pm$0.4}\end{tabular}
& \begin{tabular}[c]{@{}c@{}}46.2\\{\scriptsize$\pm$0.6}\end{tabular}
& \begin{tabular}[c]{@{}c@{}}42.1\\{\scriptsize$\pm$1.4}\end{tabular}
& \begin{tabular}[c]{@{}c@{}}72.6\\{\scriptsize$\pm$0.9}\end{tabular}
& \begin{tabular}[c]{@{}c@{}}68.8\\{\scriptsize$\pm$0.2}\end{tabular} \\
\midrule
Synth - XL-LoRA  
& \begin{tabular}[c]{@{}c@{}}81.0\\{\scriptsize$\pm$0.2}\end{tabular}
& \begin{tabular}[c]{@{}c@{}}\textbf{78.0}\\{\scriptsize$\pm$0.2}\end{tabular}
& \begin{tabular}[c]{@{}c@{}}84.3\\{\scriptsize$\pm$0.2}\end{tabular}
& \begin{tabular}[c]{@{}c@{}}\textbf{84.0}\\{\scriptsize$\pm$0.2}\end{tabular}
& \begin{tabular}[c]{@{}c@{}}\textbf{47.8}\\{\scriptsize$\pm$0.3}\end{tabular}
& \begin{tabular}[c]{@{}c@{}}\textbf{58.4}\\{\scriptsize$\pm$0.3}\end{tabular}
& \begin{tabular}[c]{@{}c@{}}72.8\\{\scriptsize$\pm$0.3}\end{tabular}
& \begin{tabular}[c]{@{}c@{}}\textbf{72.3}\\{\scriptsize$\pm$0.1}\end{tabular}
& \begin{tabular}[c]{@{}c@{}}80.4\\{\scriptsize$\pm$0.3}\end{tabular}
& \begin{tabular}[c]{@{}c@{}}\textbf{79.5}\\{\scriptsize$\pm$0.4}\end{tabular}
& \begin{tabular}[c]{@{}c@{}}\textbf{84.6}\\{\scriptsize$\pm$0.3}\end{tabular}
& \begin{tabular}[c]{@{}c@{}}\textbf{83.5}\\{\scriptsize$\pm$0.3}\end{tabular}
& \begin{tabular}[c]{@{}c@{}}49.1\\{\scriptsize$\pm$0.6}\end{tabular}
& \begin{tabular}[c]{@{}c@{}}\textbf{56.1}\\{\scriptsize$\pm$1.5}\end{tabular}
& \begin{tabular}[c]{@{}c@{}}72.6\\{\scriptsize$\pm$0.3}\end{tabular}
& \begin{tabular}[c]{@{}c@{}}\textbf{72.3}\\{\scriptsize$\pm$0.3}\end{tabular} \\
\bottomrule
\end{tabular}%
}
\caption{Embedding performance on STS tasks (Spearman’s correlation). We report results for two backbones, with mean $\pm$ standard deviation over four random seeds where applicable.}
\label{tab:str_multilingual}
\end{table*}

\begin{figure*}[tbp]
  \centering
  \includegraphics[width=\textwidth]{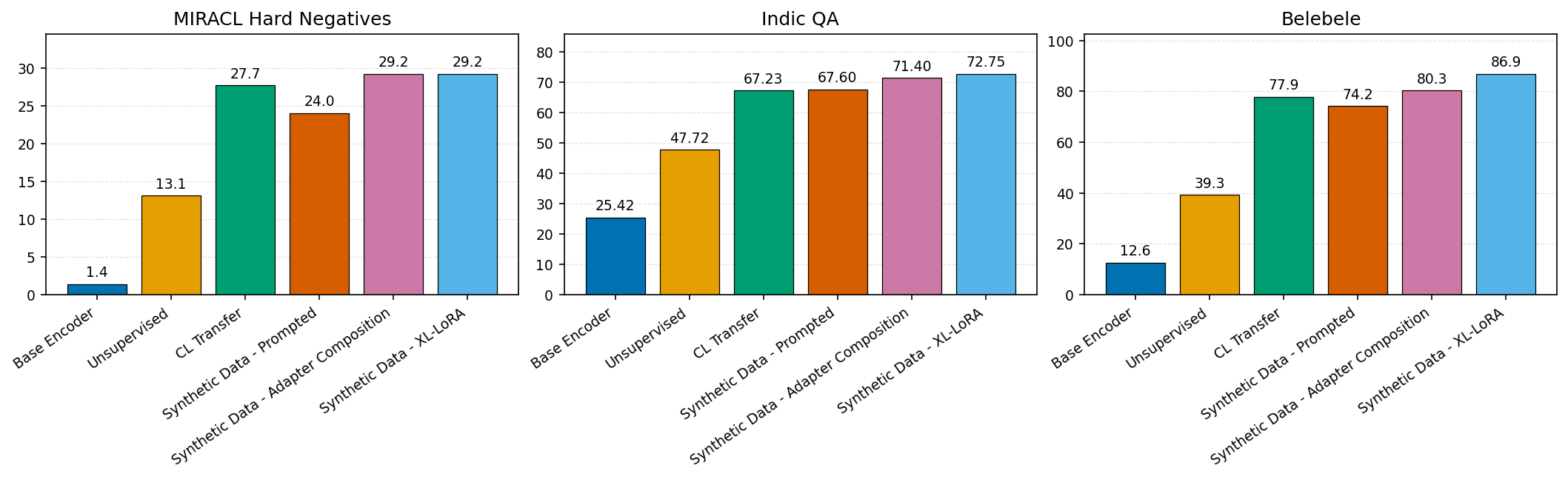}
  \caption{Retrieval results across multiple benchmarks. Results are averaged across backbones (XLM-R and mmBERT) and across languages. Metric is recall@10. Full results for all tasks and backbones can be found in Appendix~\ref{subsec:retrieval_eval}, but reflect the same clear trend depicted here.}
  \label{fig:retrieval_plot}
\end{figure*}

\subsection{Results}
STS Results can be found in Table~\ref{tab:str_multilingual}. Overall, we see that all three synthetic data approaches outperform both the base encoder and unsupervised baselines by a substantial margin. 
The cross lingual baseline nevertheless proves a tough competitor, and outperforms the prompt based synthetic approach across the board. 
However, the more sophisticated synthetic approaches, adapter composition and XL-LoRA, show consistent improvements over the cross lingual baseline, with XL-LoRA performing best of all.  
The same pattern can be seen in the retrieval results shown in Figure~\ref{fig:retrieval_plot}, where adapter composition and XL-LoRA outperform all baselines by a considerable margin, while the prompt based approach lags behind the cross lingual baseline. 
This trend is observable across backbones and across tasks, indicating that - provided the right considerations are taken in training the generator - synthetic data can be a powerful tool to bootstrap training where resources are otherwise unavailable. 
However, its success requires careful consideration and can be subject to multiple challenges and potential failure modes. In the following section, we will explore the behaviour of the various data generation methods and attempt to shed light on their strengths and weaknesses, together with their subsequent effects on the representation space. 

\section{Analysis and Ablations}
Successfully generating high quality synthetic data can be challenging, particularly in somewhat OOD settings such as low resource languages.
Here we highlight some of the key failure modes and challenges we encountered in the hope they will prove useful for future development. \\

\begin{figure}[tbp]
    \centering
    \includegraphics[width=0.5\textwidth]{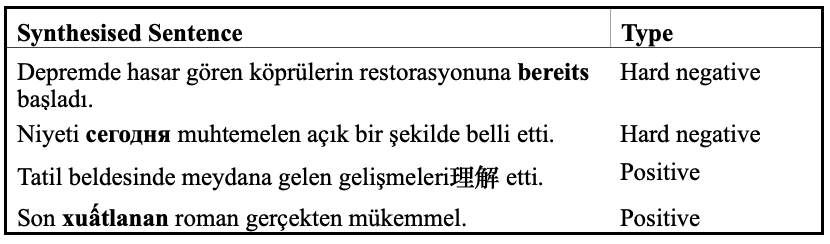}
    \caption{Synthetic Turkish data produced via ICL-Prompting. Unintended code switching and poor intellegibility are apparent throughout.}
    \label{fig:turkish_triplet_mixed_lang}
\end{figure}

\begin{figure}[htbp]
\centering
\includegraphics[width=0.45\textwidth]{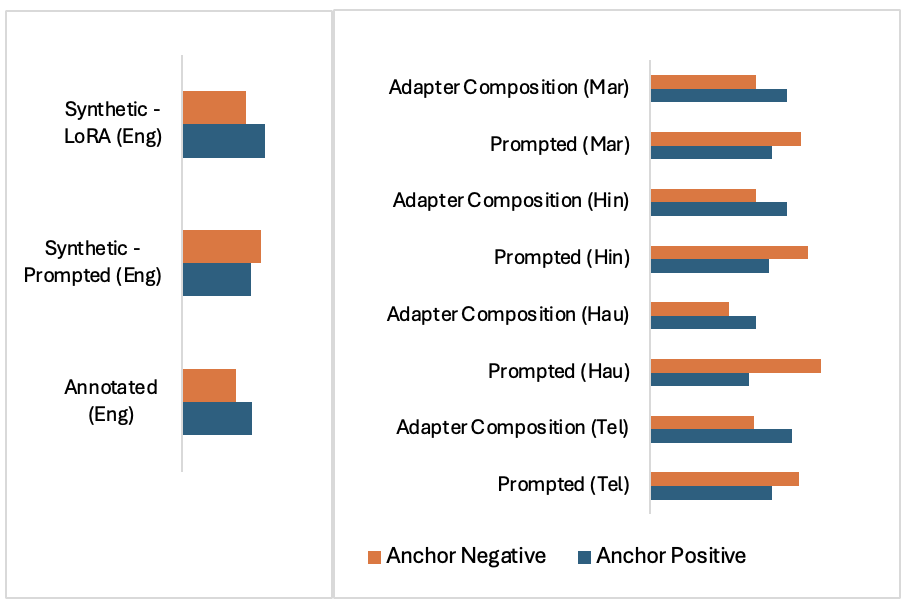}
\caption{Lexical overlap between the anchor and positive/negative pairs computed via Dice coefficient. English results, including gold human annotations, are shown on the left. Right compares adapter composition with prompted data in low resource languages. XL-LoRA is excluded as languages differ within the triplet.}
\label{fig:lexical_overlap}
\end{figure}

\noindent The first approach we tried was prompt based generation using in-context examples. This is the SynCSE approach advocated by \citet{syncse}, which showed very strong performance on standard English benchmarks, and the promise of being easily extensible to other languages and domains.
However, in our experiments we found this approach suffers from three key difficulties: 

\begin{enumerate}
  \item Lack of competency in the target language: The model often struggled to generate fluent and coherent sentences in the target language. Frequently mixing in words from other languages, and producing ungrammatical sentences. These issues compound the more low resource the language is as demonstrated in Figure~\ref{fig:turkish_triplet_mixed_lang}. 
  \item Contextual and pragmatic mismatch: The generated hard negatives often failed to be contextually or pragmatically aligned with the anchor sentence. For example, defaulting to a generic style whereas the anchor may be e.g. more conversational/informal as observed in the first example of Table~\ref{tab:triplet_qualitative}. Human annotations, meanwhile, mirror the tone. 
  \item High lexical overlap between anchor and hard negative compared to anchor and positive (Figure~\ref{fig:lexical_overlap}): While it may not be initially obvious why this is an issue, it differs strongly from the human annotations, and is indicative of the model using lazy strategies such as simply including a negation (e.g., that was \textit{not} good) which offer limited semantic diversity and scope for learning. This pattern is also observed in the qualitative analysis in Appendix~\ref{subsec:qualitative_turkish}. 
\end{enumerate}

\begin{table*}[ht]
\centering
\scriptsize
\setlength{\tabcolsep}{3pt}
\renewcommand{\arraystretch}{0.9}
\begin{tabular}{p{1.4cm} p{2.3cm} p{3.5cm} p{4.3cm} p{3.8cm}}
\toprule
& \textbf{Data type} & \textbf{Anchor} & \textbf{Positive} & \textbf{Hard Negative} \\
\midrule

\multirow{5}{*}{\textbf{Example 1}}
& Annotated
& \multirow{5}{*}{\parbox{3.5cm}{yeah it’s really sad i don’t know i just think um}}
& That’s a sad thing.
& Its an uplifting topic to talk about. \\
\cmidrule(lr){2-2}\cmidrule(lr){4-5}

& Synthetic - Prompted
&
& The person appears to be expressing emotional distress and uncertainty.
& yeah it’s great I’m sure I just feel like yeah \\
\cmidrule(lr){2-2}\cmidrule(lr){4-5}

& Synthetic - LoRA
&
& Yes, I agree that is really sad.
& Yeah I know, it’s actually very happy. \\
\midrule

\multirow{10}{*}{\textbf{Example 2}}
& Annotated
& \multirow{10}{*}{\parbox{3.5cm}{
that one person has total control and i always figured at least in a day care center there are other people around and if you get one bad apple there’s are at least other people that can see it they can watch and i just kind of always felt that the chances of something happening were less
}}
& It seems like even though there are some bad kids in daycare, there’s always someone around to supervise.
& I’ve never experienced any kids with bad behaviour in daycare. \\
\cmidrule(lr){2-2}\cmidrule(lr){4-5}

& Synthetic - Prompted
&
& I figured a daycare center with multiple staff would be safer because others can monitor and prevent bad behavior.
& I figured one bad apple could cause harm with no others to supervise. \\
\cmidrule(lr){2-2}\cmidrule(lr){4-5}

& Synthetic - LoRA
&
& I always thought that there were less chances of something happening in a day care because there are other people watching.
& I thought the chances of something happening were high. \\
\bottomrule
\end{tabular}
\caption{Synthetic triplet examples vs annotations (English).}
\label{tab:triplet_qualitative}
\end{table*}




\begin{figure}[h]
    \centering
    \includegraphics[width=0.9\linewidth]{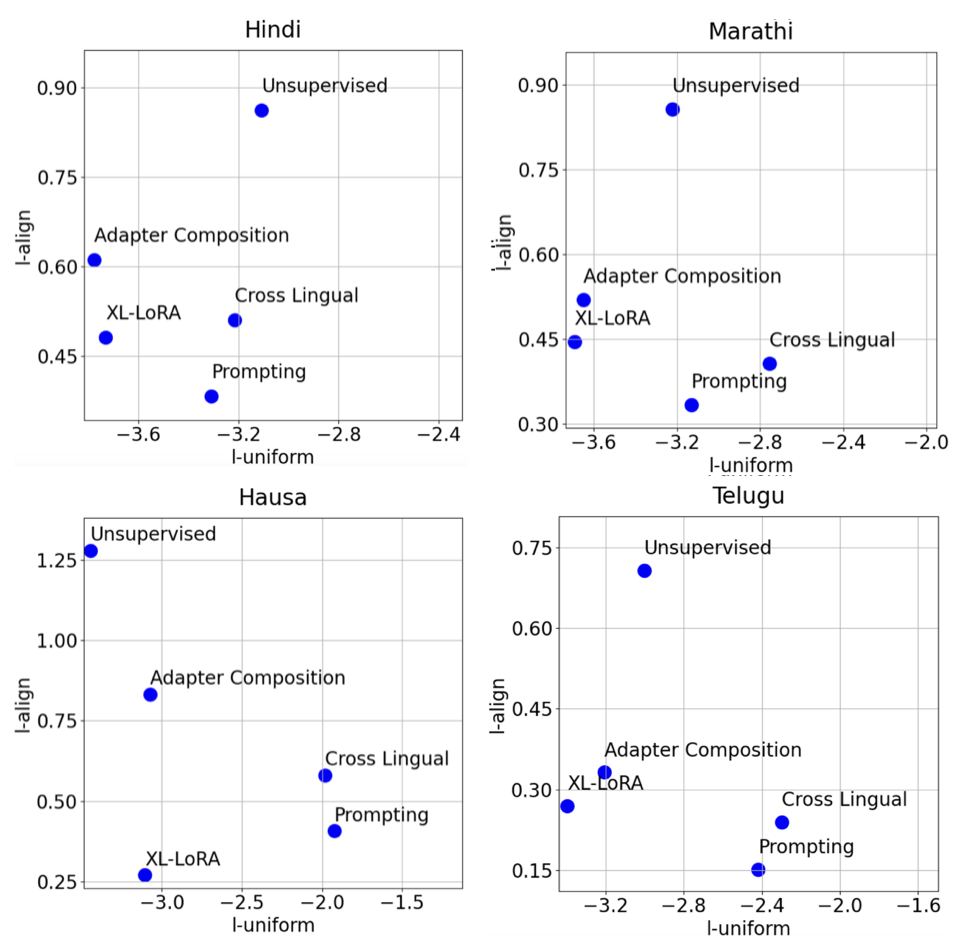}
    \caption{Alignment and Uniformity values of finetuned embedding models, based on the corresponding STR language embeddings\footnotemark. Lower $\ell_{\text{align}}$ values indicate that similar sentences are clustered closely together in the embedding space, while lower $\ell_{\text{uniform}}$ values indicate a more evenly distributed embedding space, reducing anisotropy. Here mmBERT is used as the base encoder, but trends are consistent across backbones (see ~\ref{subsec:alig_unif_tel})}
    \label{fig:lalign_lunif_all}
\end{figure}

\noindent Overall, the adapter composition approach yields stronger results than the prompting approach and shows consistent, though moderate, improvements over the cross-lingual baseline. However, while we found that task adaptation worked extremely well as seen in Appendix~\ref{subsec:english_experiments}, the performance of the composed adapters was more variable. Alignment and uniformity analysis \cite{wang2020hypersphere} in Figure~\ref{fig:lalign_lunif_all} reveals that embeddings produced by adapter composition exhibit weaker alignment than those from cross-lingual and prompting methods. This behavior is intuitive when considering that adapter composition relies on anchor–positive pairs with higher lexical overlap than the prompting synthesised method (see Figure~\ref{fig:lexical_overlap}) reducing their lexical diversity. As a result, the model may struggle to group semantically similar sentences that differ lexically. While adapter composition demonstrates a strong potential, its weaker alignment highlights clear opportunities for further refinement to enhance its efficacy in matching semantically similar sentences. \\

\begin{figure*}[ht]
\centering
\includegraphics[width=1\textwidth]{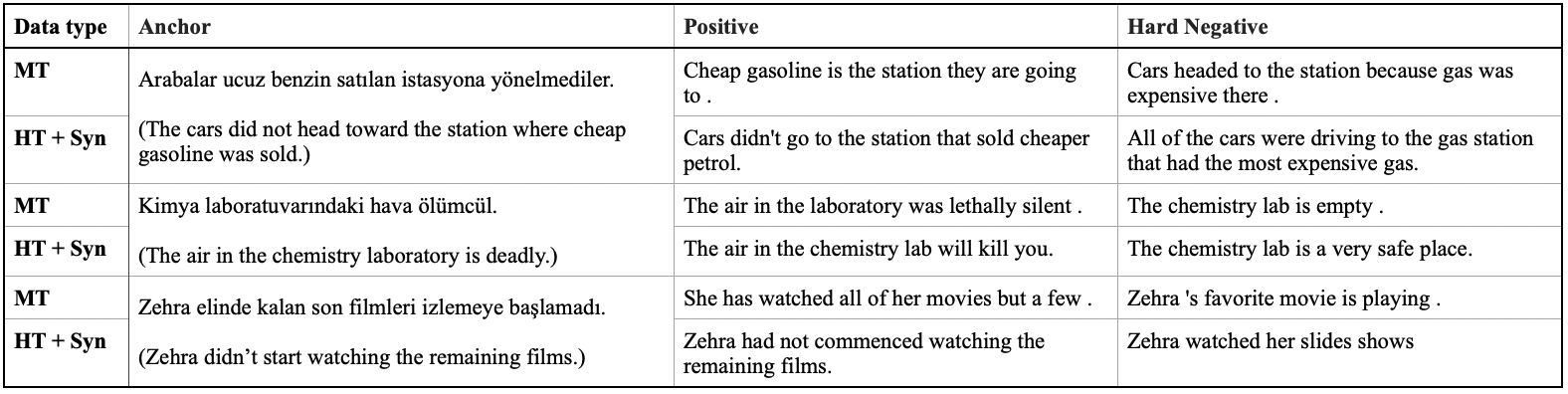}
\caption{Examples of data synthesized by the XL-LoRA generator when trained on machine-translated (MT) and human-translated plus synthesised (HT+Syn) data. MT data shows leads to lower quality.}
\label{fig:cllora_qualitative_ablation}
\end{figure*}

\footnotetext{\scriptsize Alignment and uniformity are computed per language from the STS test set, using sentence pairs with normalized similarity $\geq 0.8$ as positives and all test sentences to estimate the data distribution. }

\noindent The final model we evaluated was XL-LoRA, which leverages the stronger proficiency of models in high resource languages by generating hard positives and negatives in English while maintaining the language-specific anchors. Beyond achieving strong performance across all evaluation settings, XL-LoRA generally exhibits improved uniformity across languages relative to other synthetic approaches and cross lingual method. Furthermore, it maintains strong alignment compared to both adapter composition and unsupervised methods, as illustrated in Figure~\ref{fig:lalign_lunif_all}. This results in a more balanced embedding space that effectively clusters semantically similar pairs while preserving separation between distinct examples. \\

\noindent We further observe that XL-LoRA is highly sensitive to training data quality: when trained on XNLI machine-translated data, the LLM generator produces positives which exhibit notable semantic mismatches with their anchors, as shown in the qualitative analysis in Figure~\ref{fig:cllora_qualitative_ablation} in Appendix~\ref{subsec:cllora_ablation}. Replacing machine-translated data with XNLI human-translated data augmented by synthesised examples resolves this issue, highlighting the importance of high-quality training data for effective sentence generation.\\




\begin{table}[t]
\centering
\footnotesize
\setlength{\tabcolsep}{5pt}
\renewcommand{\arraystretch}{1.05}
\begin{tabular}{lcc}
\toprule
 & 10k & 20k \\
\midrule
hin & 79.5 $\pm$ 0.4 & \textbf{80.0 $\pm$ 0.1} \\
mar & \textbf{84.6 $\pm$ 0.3} & 83.9 $\pm$ 0.1 \\
tel & 83.5 $\pm$ 0.3 & \textbf{84.9 $\pm$ 0.6} \\
hau & 56.1 $\pm$ 1.5 & \textbf{59.3 $\pm$ 0.9} \\
\midrule
avg & 75.9 $\pm$ 0.5 & \textbf{77.0 $\pm$ 0.2} \\
\bottomrule
\end{tabular}
\caption{STS results (mean $\pm$ std over 4 seeds, Spearman's correlation) for XL-LoRA method (mmBERT as the base encoder) trained with 10k vs.\ 20k data.}
\label{tab:sts_lora_pm}
\end{table}

\noindent Finally, we conducted a limited experiment to study whether scaling the amount of training data improves the performance of the XL-LoRA adapter. In this setup, the training data for both the XL-LoRA adapter and the LoRA based task adapter used for the data augmentation step is increased from 10k to 20k. As shown in Table~\ref{tab:sts_lora_pm}, even this basic increase leads to improved STS performance across most languages. We note that this is only one axis of scaling, and there are many options including increasing source dataset diversity and LoRA rank. \\

\noindent Additional ablation studies are included in the Appendix section, detailing hyperparameter tuning for non-English sentence embedding models (Appendix~\ref{subsec:noneng-ablation}), adapter composition ablations for triplet data synthesis strategies (Appendix~\ref{subsec:adamergex-ablations}) and XL-LoRA triplet data synthesis strategies (Appendix~\ref{subsec:cllora_ablation}).  Full retrieval evaluation tables are also provided in Appendix~\ref{subsec:retrieval_eval}.

\section{Conclusions and Future Work}
We investigate whether synthetic data can be used in order to train capable embedding models for low resource languages. We find that while in-context learning through prompting is unlikely to be sufficient, more sophisticated approaches which use resource efficient LoRA adaptation of the LLM data synthesiser can prove highly effective. Notably, our best performing method, XL-LoRA, achieves considerable gains without requiring parallel data in the target language. Though we limited the number of examples for finetuning in order to maintain parity with the adapter composition. \textit{Many more} examples can be sourced both to optimise the English triplet generator, and as quality substitutions for the English anchor. With greater scale, performance should improve even further, as hinted by the results in \ref{tab:sts_lora_pm}. Compounded with the unrelenting advances in LLM capabilities, we believe this offers a bright future for low resource NLP. 

\section{Limitations}
While we aim for broad experimental coverage, extending evaluation to additional language families and analysing performance with respect to typological features such as morphology, word order, and script would be valuable future work. Our experiments are also constrained by compute, limiting evaluation to LLMs below 100B parameters; scaling to larger or more recent models may further improve synthetic data quality, particularly for lower-resource and typologically distant languages. Finally, our pipeline focuses on encoder-based embedding models, and we do not explore alternative embedding paradigms such as decoder-based embeddings from instruction-tuned LLMs, which may prove far more capable backbones. 

\section{Acknowledgments}
We thank Mirella Lapata, Edoardo Ponti, Sahil Verma and Vivek Iyer for their insightful feedback and discussions over the course of this project. We also give particular thanks to Su Kara for her valuable comments and suggestions on draft versions of this paper. Finally, MO was funded by a PhD studentship through Huawei-Edinburgh Research Lab Project 10410153 which helped to enable this work.

\clearpage
\bibliography{custom}

@inproceedings{simcse,
    title = "{S}im{CSE}: Simple Contrastive Learning of Sentence Embeddings",
    author = "Gao, Tianyu  and
      Yao, Xingcheng  and
      Chen, Danqi",
    editor = "Moens, Marie-Francine  and
      Huang, Xuanjing  and
      Specia, Lucia  and
      Yih, Scott Wen-tau",
    booktitle = "Proceedings of the 2021 Conference on Empirical Methods in Natural Language Processing",
    month = nov,
    year = "2021",
    address = "Online and Punta Cana, Dominican Republic",
    publisher = "Association for Computational Linguistics",
    url = "https://aclanthology.org/2021.emnlp-main.552/",
    doi = "10.18653/v1/2021.emnlp-main.552",
    pages = "6894--6910"
}

@misc{syncse,
      title={Contrastive Learning of Sentence Embeddings from Scratch}, 
      author={Junlei Zhang and Zhenzhong Lan and Junxian He},
      year={2023},
      eprint={2305.15077},
      archivePrefix={arXiv},
      primaryClass={cs.CL},
      url={https://arxiv.org/abs/2305.15077}, 
}

@inproceedings{sbert,
 author = {Reimers, Nils and Gurevych, Iryna},
 booktitle = emnlp-ijcnlp,
 pages = {3982--3992},
 title = {Sentence-{BERT}: Sentence Embeddings using {S}iamese {BERT}-Networks},
 year = {2019}
}

@inproceedings{devlin-etal-2019-bert,
    title = "{BERT}: Pre-training of Deep Bidirectional Transformers for Language Understanding",
    author = "Devlin, Jacob  and
      Chang, Ming-Wei  and
      Lee, Kenton  and
      Toutanova, Kristina",
    editor = "Burstein, Jill  and
      Doran, Christy  and
      Solorio, Thamar",
    booktitle = "Proceedings of the 2019 Conference of the North {A}merican Chapter of the Association for Computational Linguistics: Human Language Technologies, Volume 1 (Long and Short Papers)",
    month = jun,
    year = "2019",
    address = "Minneapolis, Minnesota",
    publisher = "Association for Computational Linguistics",
    url = "https://aclanthology.org/N19-1423/",
    doi = "10.18653/v1/N19-1423",
    pages = "4171--4186"
}

@article{bert-whitening,
  author       = {Jianlin Su and
                  Jiarun Cao and
                  Weijie Liu and
                  Yangyiwen Ou},
  title        = {Whitening Sentence Representations for Better Semantics and Faster
                  Retrieval},
  journal      = {CoRR},
  volume       = {abs/2103.15316},
  year         = {2021},
  url          = {https://arxiv.org/abs/2103.15316},
  eprinttype    = {arXiv},
  eprint       = {2103.15316},
  bibsource    = {dblp computer science bibliography, https://dblp.org}
}

@inproceedings{trafo-anisotropy,
    title = "Anisotropy is Not Inherent to Transformers",
    author = "Machina, Anemily  and
      Mercer, Robert",
    editor = "Duh, Kevin  and
      Gomez, Helena  and
      Bethard, Steven",
    booktitle = "Proceedings of the 2024 Conference of the North American Chapter of the Association for Computational Linguistics: Human Language Technologies (Volume 1: Long Papers)",
    month = jun,
    year = "2024",
    address = "Mexico City, Mexico",
    publisher = "Association for Computational Linguistics",
    url = "https://aclanthology.org/2024.naacl-long.274/",
    doi = "10.18653/v1/2024.naacl-long.274",
    pages = "4892--4907"
}

@article{ua,
  author       = {Tongzhou Wang and
                  Phillip Isola},
  title        = {Understanding Contrastive Representation Learning through Alignment
                  and Uniformity on the Hypersphere},
  journal      = {CoRR},
  volume       = {abs/2005.10242},
  year         = {2020},
  url          = {https://arxiv.org/abs/2005.10242},
  eprinttype    = {arXiv},
  eprint       = {2005.10242},
  bibsource    = {dblp computer science bibliography, https://dblp.org}
}

@inproceedings{semreltask,
title = {{S}em{E}val-2024 Task 1: Semantic Textual Relatedness for African and Asian Languages},
author = {Ousidhoum, Nedjma and Muhammad, Shamsuddeen Hassan and Abdalla, Mohamed and Abdulmumin, Idris and Ahmad,Ibrahim Said and Ahuja, Sanchit and Aji, Alham Fikri and Araujo, Vladimir and  Beloucif, Meriem and De Kock, Christine and Hourrane, Oumaima and Shrivastava, Manish and Solorio, Thamar and Surange, Nirmal and Vishnubhotla, Krishnapriya and Yimam, Seid Muhie and Mohammad, Saif M.},
booktitle = {Proceedings of the 18th International Workshop on Semantic Evaluation (SemEval-2024)},
year={2024}, 
publisher={Association for Computational Linguistics}
}

@inproceedings{bestgen-2024-satlab,
    title = "{SATL}ab at {S}em{E}val-2024 Task 1: A Fully Instance-Specific Approach for Semantic Textual Relatedness Prediction",
    author = "Bestgen, Yves",
    editor = {Ojha, Atul Kr.  and
      Do{\u{g}}ru{\"o}z, A. Seza  and
      Tayyar Madabushi, Harish  and
      Da San Martino, Giovanni  and
      Rosenthal, Sara  and
      Ros{\'a}, Aiala},
    booktitle = "Proceedings of the 18th International Workshop on Semantic Evaluation (SemEval-2024)",
    month = jun,
    year = "2024",
    address = "Mexico City, Mexico",
    publisher = "Association for Computational Linguistics",
    url = "https://aclanthology.org/2024.semeval-1.16/",
    doi = "10.18653/v1/2024.semeval-1.16",
    pages = "95--100"
}

@misc{banyan,
      title={Banyan: Improved Representation Learning with Explicit Structure}, 
      author={Mattia Opper and N. Siddharth},
      year={2025},
      eprint={2407.17771},
      archivePrefix={arXiv},
      primaryClass={cs.CL},
      url={https://arxiv.org/abs/2407.17771}, 
}

@misc{straesemeval,
      title={Self-StrAE at SemEval-2024 Task 1: Making Self-Structuring AutoEncoders Learn More With Less}, 
      author={Mattia Opper and N. Siddharth},
      year={2024},
      eprint={2404.01860},
      archivePrefix={arXiv},
      primaryClass={cs.CL},
      url={https://arxiv.org/abs/2404.01860}, 
}

@misc{thakur2021beirheterogenousbenchmarkzeroshot,
      title={BEIR: A Heterogenous Benchmark for Zero-shot Evaluation of Information Retrieval Models}, 
      author={Nandan Thakur and Nils Reimers and Andreas Rücklé and Abhishek Srivastava and Iryna Gurevych},
      year={2021},
      eprint={2104.08663},
      archivePrefix={arXiv},
      primaryClass={cs.IR},
      url={https://arxiv.org/abs/2104.08663}, 
}

@inproceedings{ni-etal-2022-sentence,
    title = "Sentence-T5: Scalable Sentence Encoders from Pre-trained Text-to-Text Models",
    author = "Ni, Jianmo  and
      Hernandez Abrego, Gustavo  and
      Constant, Noah  and
      Ma, Ji  and
      Hall, Keith  and
      Cer, Daniel  and
      Yang, Yinfei",
    editor = "Muresan, Smaranda  and
      Nakov, Preslav  and
      Villavicencio, Aline",
    booktitle = "Findings of the Association for Computational Linguistics: ACL 2022",
    month = may,
    year = "2022",
    address = "Dublin, Ireland",
    publisher = "Association for Computational Linguistics",
    url = "https://aclanthology.org/2022.findings-acl.146/",
    doi = "10.18653/v1/2022.findings-acl.146",
    pages = "1864--1874"
}

@inproceedings{wu-etal-2022-pcl,
    title = "{PCL}: Peer-Contrastive Learning with Diverse Augmentations for Unsupervised Sentence Embeddings",
    author = "Wu, Qiyu  and
      Tao, Chongyang  and
      Shen, Tao  and
      Xu, Can  and
      Geng, Xiubo  and
      Jiang, Daxin",
    editor = "Goldberg, Yoav  and
      Kozareva, Zornitsa  and
      Zhang, Yue",
    booktitle = "Proceedings of the 2022 Conference on Empirical Methods in Natural Language Processing",
    month = dec,
    year = "2022",
    address = "Abu Dhabi, United Arab Emirates",
    publisher = "Association for Computational Linguistics",
    url = "https://aclanthology.org/2022.emnlp-main.826/",
    doi = "10.18653/v1/2022.emnlp-main.826",
    pages = "12052--12066"
}

@misc{chen2020simpleframeworkcontrastivelearning,
      title={A Simple Framework for Contrastive Learning of Visual Representations}, 
      author={Ting Chen and Simon Kornblith and Mohammad Norouzi and Geoffrey Hinton},
      year={2020},
      eprint={2002.05709},
      archivePrefix={arXiv},
      primaryClass={cs.LG},
      url={https://arxiv.org/abs/2002.05709}, 
}

@inproceedings{hartvigsen-etal-2022-toxigen,
    title = "{T}oxi{G}en: A Large-Scale Machine-Generated Dataset for Adversarial and Implicit Hate Speech Detection",
    author = "Hartvigsen, Thomas  and
      Gabriel, Saadia  and
      Palangi, Hamid  and
      Sap, Maarten  and
      Ray, Dipankar  and
      Kamar, Ece",
    editor = "Muresan, Smaranda  and
      Nakov, Preslav  and
      Villavicencio, Aline",
    booktitle = "Proceedings of the 60th Annual Meeting of the Association for Computational Linguistics (Volume 1: Long Papers)",
    month = may,
    year = "2022",
    address = "Dublin, Ireland",
    publisher = "Association for Computational Linguistics",
    url = "https://aclanthology.org/2022.acl-long.234/",
    doi = "10.18653/v1/2022.acl-long.234",
    pages = "3309--3326"
}

@inproceedings{sahu-etal-2022-data,
    title = "Data Augmentation for Intent Classification with Off-the-shelf Large Language Models",
    author = "Sahu, Gaurav  and
      Rodriguez, Pau  and
      Laradji, Issam  and
      Atighehchian, Parmida  and
      Vazquez, David  and
      Bahdanau, Dzmitry",
    editor = "Liu, Bing  and
      Papangelis, Alexandros  and
      Ultes, Stefan  and
      Rastogi, Abhinav  and
      Chen, Yun-Nung  and
      Spithourakis, Georgios  and
      Nouri, Elnaz  and
      Shi, Weiyan",
    booktitle = "Proceedings of the 4th Workshop on NLP for Conversational AI",
    month = may,
    year = "2022",
    address = "Dublin, Ireland",
    publisher = "Association for Computational Linguistics",
    url = "https://aclanthology.org/2022.nlp4convai-1.5/",
    doi = "10.18653/v1/2022.nlp4convai-1.5",
    pages = "47--57"
}

@inproceedings{goldhahn-etal-2012-building,
    title = "Building Large Monolingual Dictionaries at the {L}eipzig Corpora Collection: From 100 to 200 Languages",
    author = "Goldhahn, Dirk  and
      Eckart, Thomas  and
      Quasthoff, Uwe",
    editor = "Calzolari, Nicoletta  and
      Choukri, Khalid  and
      Declerck, Thierry  and
      Do{\u{g}}an, Mehmet U{\u{g}}ur  and
      Maegaard, Bente  and
      Mariani, Joseph  and
      Moreno, Asuncion  and
      Odijk, Jan  and
      Piperidis, Stelios",
    booktitle = "Proceedings of the Eighth International Conference on Language Resources and Evaluation ({LREC}`12)",
    month = may,
    year = "2012",
    address = "Istanbul, Turkey",
    publisher = "European Language Resources Association (ELRA)",
    url = "https://aclanthology.org/L12-1154/",
    pages = "759--765"
}

@misc{conneau2020unsupervisedcrosslingualrepresentationlearning,
      title={Unsupervised Cross-lingual Representation Learning at Scale}, 
      author={Alexis Conneau and Kartikay Khandelwal and Naman Goyal and Vishrav Chaudhary and Guillaume Wenzek and Francisco Guzmán and Edouard Grave and Myle Ott and Luke Zettlemoyer and Veselin Stoyanov},
      year={2020},
      eprint={1911.02116},
      archivePrefix={arXiv},
      primaryClass={cs.CL},
      url={https://arxiv.org/abs/1911.02116}, 
}

@inproceedings{wolf-etal-2020-transformers,
    title = "Transformers: State-of-the-Art Natural Language Processing",
    author = "Wolf, Thomas  and
      Debut, Lysandre  and
      Sanh, Victor  and
      Chaumond, Julien  and
      Delangue, Clement  and
      Moi, Anthony  and
      Cistac, Pierric  and
      Rault, Tim  and
      Louf, Remi  and
      Funtowicz, Morgan  and
      Davison, Joe  and
      Shleifer, Sam  and
      von Platen, Patrick  and
      Ma, Clara  and
      Jernite, Yacine  and
      Plu, Julien  and
      Xu, Canwen  and
      Le Scao, Teven  and
      Gugger, Sylvain  and
      Drame, Mariama  and
      Lhoest, Quentin  and
      Rush, Alexander",
    editor = "Liu, Qun  and
      Schlangen, David",
    booktitle = "Proceedings of the 2020 Conference on Empirical Methods in Natural Language Processing: System Demonstrations",
    month = oct,
    year = "2020",
    address = "Online",
    publisher = "Association for Computational Linguistics",
    url = "https://aclanthology.org/2020.emnlp-demos.6/",
    doi = "10.18653/v1/2020.emnlp-demos.6",
    pages = "38--45"
}

@misc{cer2018universalsentenceencoder,
      title={Universal Sentence Encoder}, 
      author={Daniel Cer and Yinfei Yang and Sheng-yi Kong and Nan Hua and Nicole Limtiaco and Rhomni St. John and Noah Constant and Mario Guajardo-Cespedes and Steve Yuan and Chris Tar and Yun-Hsuan Sung and Brian Strope and Ray Kurzweil},
      year={2018},
      eprint={1803.11175},
      archivePrefix={arXiv},
      primaryClass={cs.CL},
      url={https://arxiv.org/abs/1803.11175}, 
}

@article{gemma_2025,
    title={Gemma 3},
    url={https://goo.gle/Gemma3Report},
    publisher={Kaggle},
    author={Gemma Team},
    year={2025}
}

@inproceedings{wang2020hypersphere,
  title     = {Understanding Contrastive Representation Learning through Alignment and Uniformity on the Hypersphere},
  author    = {Wang, Tongzhou and Isola, Phillip},
  booktitle = {International Conference on Machine Learning (ICML)},
  organization = {PMLR},
  pages     = {9929--9939},
  year      = {2020}
}

@misc{modernbert,
      title={Smarter, Better, Faster, Longer: A Modern Bidirectional Encoder for Fast, Memory Efficient, and Long Context Finetuning and Inference}, 
      author={Benjamin Warner and Antoine Chaffin and Benjamin Clavié and Orion Weller and Oskar Hallström and Said Taghadouini and Alexis Gallagher and Raja Biswas and Faisal Ladhak and Tom Aarsen and Nathan Cooper and Griffin Adams and Jeremy Howard and Iacopo Poli},
      year={2024},
      eprint={2412.13663},
      archivePrefix={arXiv},
      primaryClass={cs.CL},
      url={https://arxiv.org/abs/2412.13663}, 
}

@misc{mmbert,
      title={mmBERT: A Modern Multilingual Encoder with Annealed Language Learning}, 
      author={Marc Marone and Orion Weller and William Fleshman and Eugene Yang and Dawn Lawrie and Benjamin Van Durme},
      year={2025},
      eprint={2509.06888},
      archivePrefix={arXiv},
      primaryClass={cs.CL},
      url={https://arxiv.org/abs/2509.06888}, 
}

@inproceedings{halat-atlamaz-2024-implicatr,
    title = "{I}mplica{TR}: A Granular Dataset for Natural Language Inference and Pragmatic Reasoning in {T}urkish",
    author = {Halat, Mustafa  and
      Atlamaz, {\"U}mit},
    editor = {Ataman, Duygu  and
      Derin, Mehmet Oguz  and
      Ivanova, Sardana  and
      K{\"o}ksal, Abdullatif  and
      S{\"a}lev{\"a}, Jonne  and
      Zeyrek, Deniz},
    booktitle = "Proceedings of the First Workshop on Natural Language Processing for Turkic Languages (SIGTURK 2024)",
    month = aug,
    year = "2024",
    address = "Bangkok, Thailand and Online",
    publisher = "Association for Computational Linguistics",
    url = "https://aclanthology.org/2024.sigturk-1.3/",
    pages = "29--41"
}

@article{muennighoff2022mteb,
  author = {Muennighoff, Niklas and Tazi, Nouamane and Magne, Lo{\"\i}c and Reimers, Nils},
  title = {MTEB: Massive Text Embedding Benchmark},
  publisher = {arXiv},
  journal={arXiv preprint arXiv:2210.07316},
  year = {2022},
  url = {https://arxiv.org/abs/2210.07316},
  doi = {10.48550/ARXIV.2210.07316},
}

@inproceedings{huang-etal-2023-languages,
    title = "Not All Languages Are Created Equal in {LLM}s: Improving Multilingual Capability by Cross-Lingual-Thought Prompting",
    author = "Huang, Haoyang  and
      Tang, Tianyi  and
      Zhang, Dongdong  and
      Zhao, Xin  and
      Song, Ting  and
      Xia, Yan  and
      Wei, Furu",
    editor = "Bouamor, Houda  and
      Pino, Juan  and
      Bali, Kalika",
    booktitle = "Findings of the Association for Computational Linguistics: EMNLP 2023",
    month = dec,
    year = "2023",
    address = "Singapore",
    publisher = "Association for Computational Linguistics",
    url = "https://aclanthology.org/2023.findings-emnlp.826/",
    doi = "10.18653/v1/2023.findings-emnlp.826",
    pages = "12365--12394"
}

@article{colbert-x, 
  author       = {Suraj Nair and
                  Eugene Yang and
                  Dawn J. Lawrie and
                  Kevin Duh and
                  Paul McNamee and
                  Kenton Murray and
                  James Mayfield and
                  Douglas W. Oard},
  title        = {Transfer Learning Approaches for Building Cross-Language Dense Retrieval
                  Models},
  journal      = {CoRR},
  volume       = {abs/2201.08471},
  year         = {2022},
  url          = {https://arxiv.org/abs/2201.08471},
  eprinttype    = {arXiv},
  eprint       = {2201.08471},
  bibsource    = {dblp computer science bibliography, https://dblp.org}
}

@article{williams-etal-2018-broad,
  author       = {Adina Williams and
                  Nikita Nangia and
                  Samuel R. Bowman},
  title        = {A Broad-Coverage Challenge Corpus for Sentence Understanding through
                  Inference},
  journal      = {CoRR},
  volume       = {abs/1704.05426},
  year         = {2017},
  url          = {http://arxiv.org/abs/1704.05426},
  eprinttype    = {arXiv},
  eprint       = {1704.05426},
  bibsource    = {dblp computer science bibliography, https://dblp.org}
}

@misc{hu2021loralowrankadaptationlarge,
      title={LoRA: Low-Rank Adaptation of Large Language Models}, 
      author={Edward J. Hu and Yelong Shen and Phillip Wallis and Zeyuan Allen-Zhu and Yuanzhi Li and Shean Wang and Lu Wang and Weizhu Chen},
      year={2021},
      eprint={2106.09685},
      archivePrefix={arXiv},
      primaryClass={cs.CL},
      url={https://arxiv.org/abs/2106.09685}, 
}

@inproceedings{wang-etal-2023-self-instruct,
    title = "Self-Instruct: Aligning Language Models with Self-Generated Instructions",
    author = "Wang, Yizhong  and
      Kordi, Yeganeh  and
      Mishra, Swaroop  and
      Liu, Alisa  and
      Smith, Noah A.  and
      Khashabi, Daniel  and
      Hajishirzi, Hannaneh",
    editor = "Rogers, Anna  and
      Boyd-Graber, Jordan  and
      Okazaki, Naoaki",
    booktitle = "Proceedings of the 61st Annual Meeting of the Association for Computational Linguistics (Volume 1: Long Papers)",
    month = jul,
    year = "2023",
    address = "Toronto, Canada",
    publisher = "Association for Computational Linguistics",
    url = "https://aclanthology.org/2023.acl-long.754/",
    doi = "10.18653/v1/2023.acl-long.754",
    pages = "13484--13508"
}

@inproceedings{honovich-etal-2023-unnatural,
    title = "Unnatural Instructions: Tuning Language Models with (Almost) No Human Labor",
    author = "Honovich, Or  and
      Scialom, Thomas  and
      Levy, Omer  and
      Schick, Timo",
    editor = "Rogers, Anna  and
      Boyd-Graber, Jordan  and
      Okazaki, Naoaki",
    booktitle = "Proceedings of the 61st Annual Meeting of the Association for Computational Linguistics (Volume 1: Long Papers)",
    month = jul,
    year = "2023",
    address = "Toronto, Canada",
    publisher = "Association for Computational Linguistics",
    url = "https://aclanthology.org/2023.acl-long.806/",
    doi = "10.18653/v1/2023.acl-long.806",
    pages = "14409--14428"
}

@misc{whitehouse2024lowrankadaptationmultilingualsummarization,
      title={Low-Rank Adaptation for Multilingual Summarization: An Empirical Study}, 
      author={Chenxi Whitehouse and Fantine Huot and Jasmijn Bastings and Mostafa Dehghani and Chu-Cheng Lin and Mirella Lapata},
      year={2024},
      eprint={2311.08572},
      archivePrefix={arXiv},
      primaryClass={cs.CL},
      url={https://arxiv.org/abs/2311.08572}, 
}

@misc{zhao2024adamergexcrosslingualtransferlarge,
      title={AdaMergeX: Cross-Lingual Transfer with Large Language Models via Adaptive Adapter Merging}, 
      author={Yiran Zhao and Wenxuan Zhang and Huiming Wang and Kenji Kawaguchi and Lidong Bing},
      year={2024},
      eprint={2402.18913},
      archivePrefix={arXiv},
      primaryClass={cs.CL},
      url={https://arxiv.org/abs/2402.18913}, 
}

@misc{zhang2023composingparameterefficientmodulesarithmetic,
      title={Composing Parameter-Efficient Modules with Arithmetic Operations}, 
      author={Jinghan Zhang and Shiqi Chen and Junteng Liu and Junxian He},
      year={2023},
      eprint={2306.14870},
      archivePrefix={arXiv},
      primaryClass={cs.CL},
      url={https://arxiv.org/abs/2306.14870}, 
}

@inproceedings{bowman-etal-2015-large,
    title = "A large annotated corpus for learning natural language inference",
    author = "Bowman, Samuel R.  and
      Angeli, Gabor  and
      Potts, Christopher  and
      Manning, Christopher D.",
    editor = "M{\`a}rquez, Llu{\'i}s  and
      Callison-Burch, Chris  and
      Su, Jian",
    booktitle = "Proceedings of the 2015 Conference on Empirical Methods in Natural Language Processing",
    month = sep,
    year = "2015",
    address = "Lisbon, Portugal",
    publisher = "Association for Computational Linguistics",
    url = "https://aclanthology.org/D15-1075/",
    doi = "10.18653/v1/D15-1075",
    pages = "632--642"
}

@misc{singh2024aya,
      title={Aya Dataset: An Open-Access Collection for Multilingual Instruction Tuning}, 
      author={Shivalika Singh and Freddie Vargus and Daniel Dsouza and Börje F. Karlsson and Abinaya Mahendiran and Wei-Yin Ko and Herumb Shandilya and Jay Patel and Deividas Mataciunas and Laura OMahony and Mike Zhang and Ramith Hettiarachchi and Joseph Wilson and Marina Machado and Luisa Souza Moura and Dominik Krzemiński and Hakimeh Fadaei and Irem Ergün and Ifeoma Okoh and Aisha Alaagib and Oshan Mudannayake and Zaid Alyafeai and Vu Minh Chien and Sebastian Ruder and Surya Guthikonda and Emad A. Alghamdi and Sebastian Gehrmann and Niklas Muennighoff and Max Bartolo and Julia Kreutzer and Ahmet Üstün and Marzieh Fadaee and Sara Hooker},
      year={2024},
      eprint={2402.06619},
      archivePrefix={arXiv},
      primaryClass={cs.CL}
}

@misc{ahuja2023megamultilingualevaluationgenerative,
      title={MEGA: Multilingual Evaluation of Generative AI}, 
      author={Kabir Ahuja and Harshita Diddee and Rishav Hada and Millicent Ochieng and Krithika Ramesh and Prachi Jain and Akshay Nambi and Tanuja Ganu and Sameer Segal and Maxamed Axmed and Kalika Bali and Sunayana Sitaram},
      year={2023},
      eprint={2303.12528},
      archivePrefix={arXiv},
      primaryClass={cs.CL},
      url={https://arxiv.org/abs/2303.12528}, 
}

@misc{pomerenke2025ailanguageproficiencymonitor,
      title={The AI Language Proficiency Monitor -- Tracking the Progress of LLMs on Multilingual Benchmarks}, 
      author={David Pomerenke and Jonas Nothnagel and Simon Ostermann},
      year={2025},
      eprint={2507.08538},
      archivePrefix={arXiv},
      primaryClass={cs.CL},
      url={https://arxiv.org/abs/2507.08538}, 
}

@inproceedings{conneau-etal-2018-xnli,
    title = "{XNLI}: Evaluating Cross-lingual Sentence Representations",
    author = "Conneau, Alexis  and
      Rinott, Ruty  and
      Lample, Guillaume  and
      Williams, Adina  and
      Bowman, Samuel  and
      Schwenk, Holger  and
      Stoyanov, Veselin",
    editor = "Riloff, Ellen  and
      Chiang, David  and
      Hockenmaier, Julia  and
      Tsujii, Jun{'}ichi",
    booktitle = "Proceedings of the 2018 Conference on Empirical Methods in Natural Language Processing",
    month = oct # "-" # nov,
    year = "2018",
    address = "Brussels, Belgium",
    publisher = "Association for Computational Linguistics",
    url = "https://aclanthology.org/D18-1269/",
    doi = "10.18653/v1/D18-1269",
    pages = "2475--2485"
}

@inproceedings{tiedemann-de-gibert-2023-opus,
    title = "The {OPUS}-{MT} Dashboard {--} A Toolkit for a Systematic Evaluation of Open Machine Translation Models",
    author = {Tiedemann, J{\"o}rg  and
      de Gibert, Ona},
    editor = "Bollegala, Danushka  and
      Huang, Ruihong  and
      Ritter, Alan",
    booktitle = "Proceedings of the 61st Annual Meeting of the Association for Computational Linguistics (Volume 3: System Demonstrations)",
    month = jul,
    year = "2023",
    address = "Toronto, Canada",
    publisher = "Association for Computational Linguistics",
    url = "https://aclanthology.org/2023.acl-demo.30/",
    doi = "10.18653/v1/2023.acl-demo.30",
    pages = "315--327"
}

@article{labse,
  author       = {Fangxiaoyu Feng and
                  Yinfei Yang and
                  Daniel Cer and
                  Naveen Arivazhagan and
                  Wei Wang},
  title        = {Language-agnostic {BERT} Sentence Embedding},
  journal      = {CoRR},
  volume       = {abs/2007.01852},
  year         = {2020},
  url          = {https://arxiv.org/abs/2007.01852},
  eprinttype    = {arXiv},
  eprint       = {2007.01852},
  bibsource    = {dblp computer science bibliography, https://dblp.org}
}

@article{muse,
  title={Unsupervised Machine Translation Using Monolingual Corpora Only},
  author={Lample, Guillaume and Conneau, Alexis and Denoyer, Ludovic and Ranzato, Marc'Aurelio},
  journal={arXiv preprint arXiv:1711.00043},
  year={2017}
}

@article{parascale,
  author       = {John Wieting and
                  Kevin Gimpel and
                  Graham Neubig and
                  Taylor Berg{-}Kirkpatrick},
  title        = {Paraphrastic Representations at Scale},
  journal      = {CoRR},
  volume       = {abs/2104.15114},
  year         = {2021},
  url          = {https://arxiv.org/abs/2104.15114},
  eprinttype    = {arXiv},
  eprint       = {2104.15114},
  bibsource    = {dblp computer science bibliography, https://dblp.org}
}

@misc{laser3,
      title={Bitext Mining Using Distilled Sentence Representations for Low-Resource Languages}, 
      author={Kevin Heffernan and Onur Çelebi and Holger Schwenk},
      year={2022},
      eprint={2205.12654},
      archivePrefix={arXiv},
      primaryClass={cs.CL},
      url={https://arxiv.org/abs/2205.12654}, 
}

@misc{mao2023leallalearninglightweightlanguageagnostic,
      title={LEALLA: Learning Lightweight Language-agnostic Sentence Embeddings with Knowledge Distillation}, 
      author={Zhuoyuan Mao and Tetsuji Nakagawa},
      year={2023},
      eprint={2302.08387},
      archivePrefix={arXiv},
      primaryClass={cs.CL},
      url={https://arxiv.org/abs/2302.08387}, 
}

@inproceedings{reimers-gurevych-2020-making,
    title = "Making Monolingual Sentence Embeddings Multilingual using Knowledge Distillation",
    author = "Reimers, Nils  and
      Gurevych, Iryna",
    editor = "Webber, Bonnie  and
      Cohn, Trevor  and
      He, Yulan  and
      Liu, Yang",
    booktitle = "Proceedings of the 2020 Conference on Empirical Methods in Natural Language Processing (EMNLP)",
    month = nov,
    year = "2020",
    address = "Online",
    publisher = "Association for Computational Linguistics",
    url = "https://aclanthology.org/2020.emnlp-main.365/",
    doi = "10.18653/v1/2020.emnlp-main.365",
    pages = "4512--4525"
}

@inproceedings{nygaard2003opus,
title = {OPUS—an open source parallel corpus},
author = {Nygaard, Lars and Tiedemann, J{\"o}rg},
booktitle = {Proceedings of the 13th Nordic Conference on Computational Linguistics},
year = {2003}
}

@inproceedings{wendler-etal-2024-llamas,
    title = "Do Llamas Work in {E}nglish? On the Latent Language of Multilingual Transformers",
    author = "Wendler, Chris  and
      Veselovsky, Veniamin  and
      Monea, Giovanni  and
      West, Robert",
    editor = "Ku, Lun-Wei  and
      Martins, Andre  and
      Srikumar, Vivek",
    booktitle = "Proceedings of the 62nd Annual Meeting of the Association for Computational Linguistics (Volume 1: Long Papers)",
    month = aug,
    year = "2024",
    address = "Bangkok, Thailand",
    publisher = "Association for Computational Linguistics",
    url = "https://aclanthology.org/2024.acl-long.820/",
    doi = "10.18653/v1/2024.acl-long.820",
    pages = "15366--15394"
}

@article{rag,
  author       = {Patrick Lewis and
                  Ethan Perez and
                  Aleksandra Piktus and
                  Fabio Petroni and
                  Vladimir Karpukhin and
                  Naman Goyal and
                  Heinrich K{\"{u}}ttler and
                  Mike Lewis and
                  Wen{-}tau Yih and
                  Tim Rockt{\"{a}}schel and
                  Sebastian Riedel and
                  Douwe Kiela},
  title        = {Retrieval-Augmented Generation for Knowledge-Intensive {NLP} Tasks},
  journal      = {CoRR},
  volume       = {abs/2005.11401},
  year         = {2020},
  url          = {https://arxiv.org/abs/2005.11401},
  eprinttype    = {arXiv},
  eprint       = {2005.11401},
  bibsource    = {dblp computer science bibliography, https://dblp.org}
}

@misc{bubeck2023sparksartificialgeneralintelligence,
      title={Sparks of Artificial General Intelligence: Early experiments with GPT-4}, 
      author={Sébastien Bubeck and Varun Chandrasekaran and Ronen Eldan and Johannes Gehrke and Eric Horvitz and Ece Kamar and Peter Lee and Yin Tat Lee and Yuanzhi Li and Scott Lundberg and Harsha Nori and Hamid Palangi and Marco Tulio Ribeiro and Yi Zhang},
      year={2023},
      eprint={2303.12712},
      archivePrefix={arXiv},
      primaryClass={cs.CL},
      url={https://arxiv.org/abs/2303.12712}, 
}

\clearpage
\appendix
\section{Appendix}
\label{sec:appendix} 

\subsection{English Experiments Setup}
\label{subsec:english_experiments}

\begin{table}[H]
\centering
\scriptsize
\setlength{\tabcolsep}{4pt}
\renewcommand{\arraystretch}{0.8}

\begin{tabular}{l l l c}
\toprule
\textbf{\makecell[l]{Base\\Encoder}} &
\textbf{\makecell[l]{Base Model\\Description}} &
\textbf{\makecell[l]{Synthesised\\Data\\Generation\\Method}} &
\textbf{STS-B} \\
\midrule
\multirow[c]{15}{*}{\makecell[l]{RoBERTa\\Base}}
 & \makecell[l]{unsupervised\\SimCSE} & -- & 80.0 \\
\addlinespace[2.5pt]
 & \makecell[l]{supervised-SimCSE\\on annotated\\English triplets}
 & -- & \textbf{85.4} \\
\addlinespace[2.5pt]
 & \multirow[c]{8}{*}{\makecell[c]{supervised\\SimCSE}}
 & \makecell[l]{SynCSE-partial\\(GPT3.5)} & 83.3 \\
\addlinespace[2.5pt]
 &  
 & \makecell[l]{SynCSE-partial\\(Gemma3-27b)*} & 83.1 \\

 &  
 & \makecell[l]{LoRA TA\\(Gemma3-1b)*} & 83.2 \\

 &  
 & \makecell[l]{LoRA TA\\(Gemma3-27b)*} & 84.3 \\
\bottomrule
\end{tabular}

\caption{STS-B performance (Spearman’s correlation) of SimCSE-based English models. Results are reproduced via SimCSE fine-tuning methodologies, using their hyperparameter settings. * denotes synthesised data. Original scores for unsupervised SimCSE, supervised SimCSE, and SynCSE-partial (GPT-3.5) are 80.2, 85.8, and 83.9 \citep{simcse, syncse}.} 
\label{tab:sts_b_roberta}
\end{table}

\subsection{Non-English Setup Ablations}
\label{subsec:noneng-ablation}

\subsubsection{Sentence Embedding Models Training and Hyperparameter Settings}
\label{subsubsec:training_details}

To train our models, we utilise the SimCSE package, which is built on top of the Transformers library \cite{wolf-etal-2020-transformers}, modifying the package to be able to extend the base encoder use to ModernBERT \citep{modernbert} architecture. Our approach builds upon pre-trained language models, leveraging the XLM-RoBERTa Base model \cite{conneau2020unsupervisedcrosslingualrepresentationlearning} and the mmBERT Base model \cite{mmbert} as the foundations for our low resource language models. We drew upon the hyperparameters established in the original SynCSE and SimCSE studies except the pooling method. Our ablation study indicates that using the average first last pooling method during training and inference, a batch size of 512, a learning rate of 5e-5 and a maximum sequence length of 32 consistently results in high scores across two different languages. For training the unsupervised SimCSE models, we use the same hyperparameters as the original unsupervised SimCSE, including a batch size of 512, a maximum sequence length of 32, and a learning rate of 1e-5. The ablation studies were conducted on the development sets of the evaluation data. All results report the average first–last pooling method, except for the Base Encoder models in the STS experiments, which use average pooling, as this pooler yields higher baseline scores for this specific task.

\begin{table}[h]
\centering
\scriptsize
\renewcommand{\arraystretch}{0.9}
\begin{tabular}{@{}lcc@{}}
\toprule
\textbf{Pooler Type} & \multicolumn{2}{c}{\textbf{STR}} \\
\cmidrule(lr){2-3}
 & \textbf{Afr} & \textbf{Hin} \\
\hline
cls + mlp & 77.9        & 81.8 \\
avg       & 77.6        & 81.3 \\
afl       & \textbf{78.4} & \textbf{82.0} \\
at2       & 78.0        & 81.6 \\
cbp       & 77.7        & 81.8 \\
\bottomrule
\end{tabular}
\caption{Spearman correlations for STR performance fine-tuned with Prompting synthesis method in Afrikaans and Hindi, using different pooler types}
\label{table:ablation-pooler}
\end{table}

\begin{table}[h]
\centering
\scriptsize
\renewcommand{\arraystretch}{0.9}
\begin{tabular}{@{}cccc@{}}
\toprule
\textbf{\makecell[tl]{Batch\\Size}} & 
\textbf{\makecell[tl]{MLM\\Obj}} & 
\multicolumn{2}{c}{\textbf{STR}} \\
\cmidrule(lr){3-4}
 & & \textbf{Afr} & \textbf{Hin} \\
\hline
128 & FALSE & 78.4       & 82.0 \\
128 & TRUE  & 78.5       & 82.1 \\
512 & FALSE & \textbf{79.0} & \textbf{82.8} \\
\bottomrule
\end{tabular}
\caption{Spearman correlations for STR performance fine-tuned with Prompting synthesis method in Afrikaans and Hindi, using different batch sizes and MLM objective.}
\label{table:ablation-mlmbatch}
\end{table}

\subsection{Localised SynCSE-partial pipeline}
\label{subsec:localised-syncse-partial}


\begin{figure}[H]
  \centering
  \includegraphics[width=0.6\textwidth]{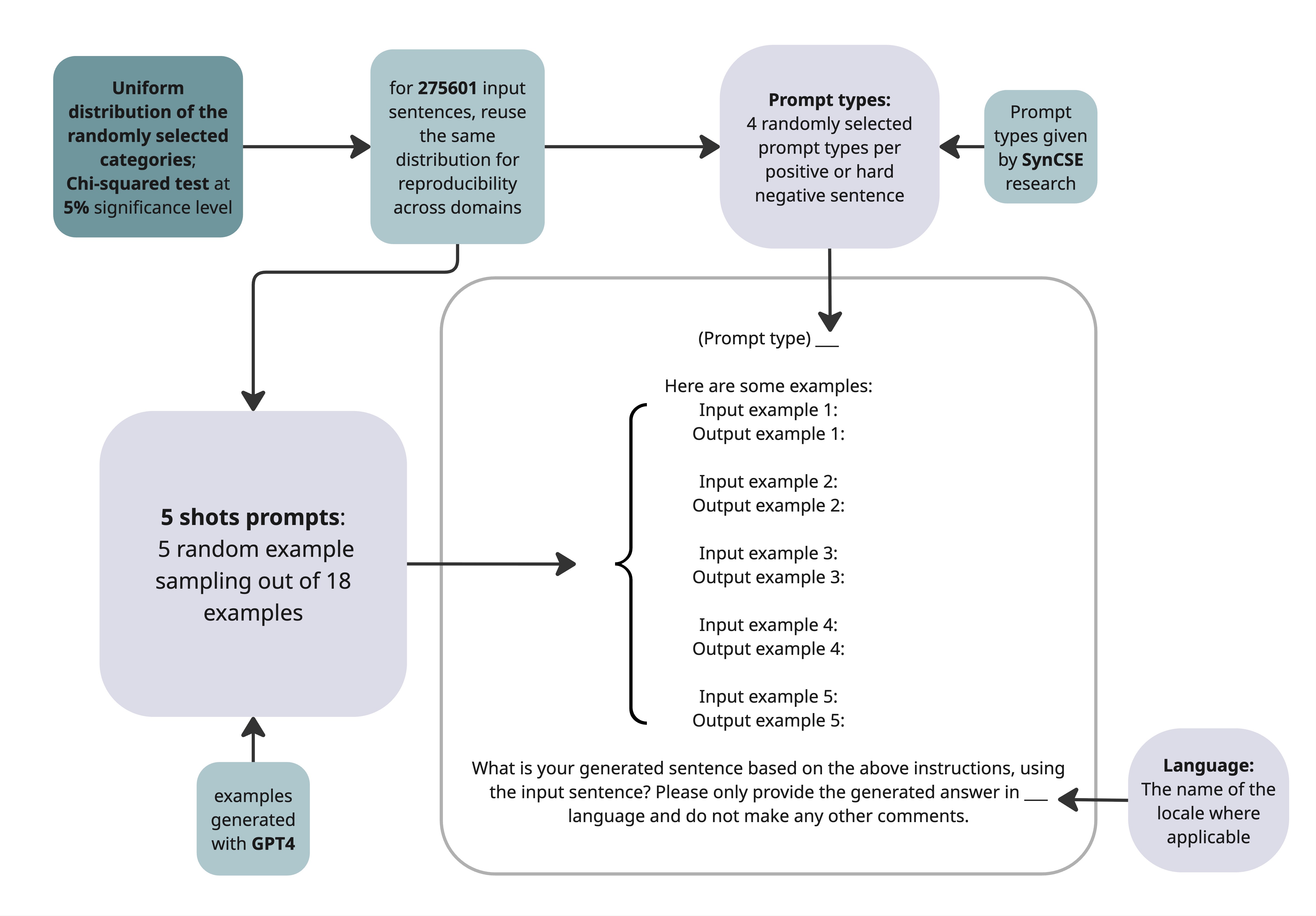}
  \caption{The prompting strategy for localised SynCSE-partial. As in the original method, each prompt includes five randomly selected examples from a pool of 18, together with a randomly chosen opening sentence from one of the four prompt types shown in Figure~\ref{fig:prompt_pools}. Language specific details are added in the closing prompt.}
  \label{fig:prompting-strategy}
\end{figure}

\subsection{SynCSE-partial Positive and Hard Negative Prompts}
\label{subsec:SynCSE-prompts}

\begin{figure}[H]
\centering
\includegraphics[width=0.5\textwidth]{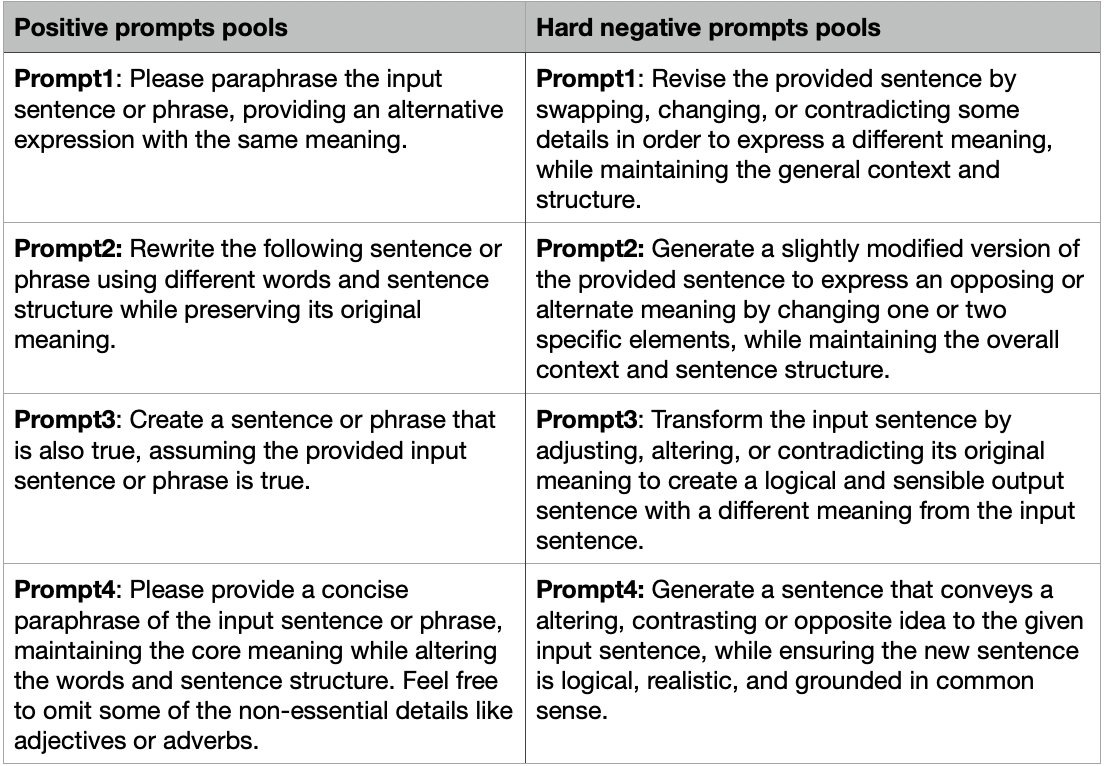}
\caption{Positive and hard negative prompts used in the SynCSE-partial methodology \citep{syncse} to promote diversity in data synthesis.}
\label{fig:prompt_pools}
\end{figure}

\subsection{Adapter Composition Triplet Data Synthesis Ablations}
\label{subsec:adamergex-ablations}

\subsubsection{Ablation Design}
\begin{figure}[H]
\centering
\includegraphics[width=0.45\textwidth]{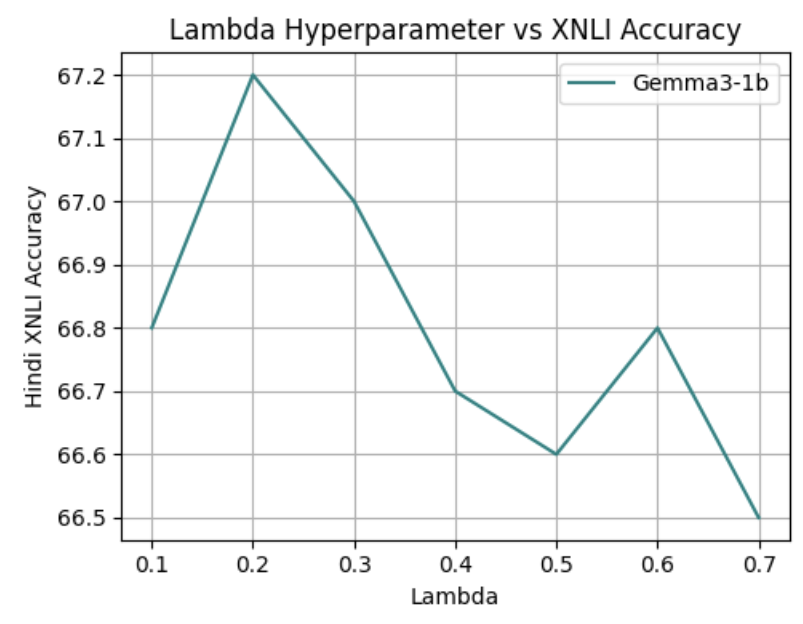}\\[0.5em]
\includegraphics[width=0.5\textwidth]{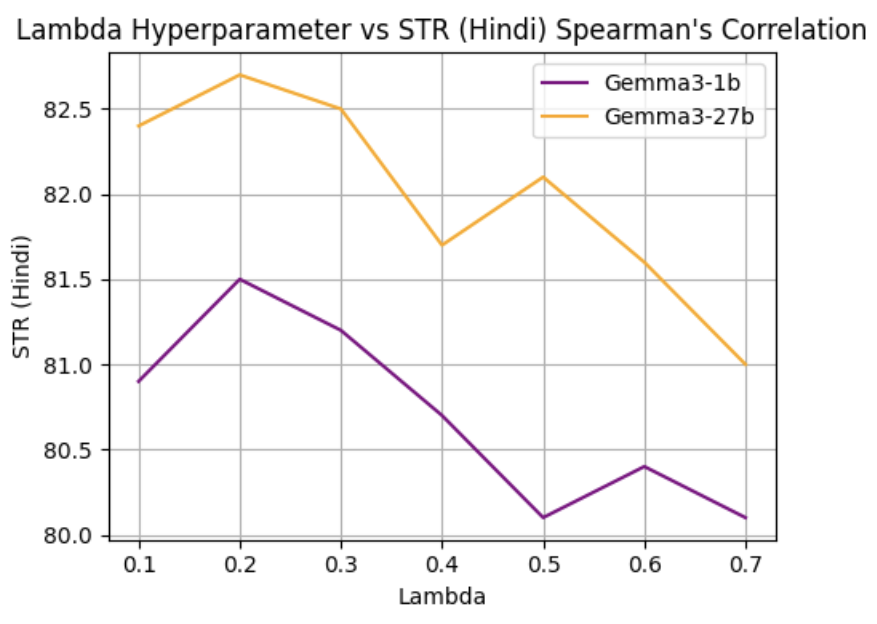}
\caption{Accuracy for XNLI (dev) and STR (dev) performances across different lambdas.}
\label{fig:lambda_param}
\end{figure}

Ablation studies of the adapter composition method for triplet data synthesis include reproducing the pipeline of  \cite{zhao2024adamergexcrosslingualtransferlarge} for the XNLI task using Gemma3 1B model and implementing the approach to the STR task for Gemma3 1B and 27B models. For ensuring the effectiveness of the implementation, we tested the performance of different task adapter training methods (Table~\ref{tab:str_hindi_groups}) and different sources of training data for the language adapters (Table~\ref{tab:str_grouped}), rank values (Table~\ref{tab:str_ranks}) and lambda hyperparameters (Figure~\ref{fig:lambda_param}). The development set is used for the STR evaluation throughout the ablation experiments. The initial experiments synthesise a smaller set (50k) of data due to compute limits.

\subsubsection{LoRA Hyperparameters}
For training all the LoRA adapters, the following hyperparameters are used; \text{lora\_alpha}: 16, \text{target\_modules}: all-linear following the AdaMergeX parameters. Rank is set to 8 based on our ablation results.

\begin{table}[h]
\centering
\scriptsize
\setlength{\tabcolsep}{6pt}
\renewcommand{\arraystretch}{0.9}

\begin{tabular}{l l cc}
\toprule
\textbf{Task Adapter} & \textbf{\makecell[l]{Adapter\\Composition\\Method}} & \multicolumn{2}{c}{\textbf{XNLI Accuracy Score}} \\
\cmidrule(l){3-4}
 &  & \textbf{XLM-R} & \textbf{Gemma3-1b} \\
\midrule
Base model           & --                                                        & 33.3 & 35.3 \\
Hindi XNLI   & --                                                       & 57.2 & 65.7 \\
English XNLI & --                                                       & 56.8 & 65.3 \\
English XNLI & \makecell[l]{AdaMergeX with \\Hindi LA \& English LA}    & \textbf{60.5} & \textbf{65.8} \\
\bottomrule
\end{tabular}%
\caption{XNLI accuracy on the Hindi test set for Base models, LoRA task adapters (TAs) in Hindi and English, and the AdaMergeX cross-lingual TA.}
\label{tab:xnli_adapters}
\end{table}

\begin{table}[H]
\centering
\scriptsize
\setlength{\tabcolsep}{6pt}
\renewcommand{\arraystretch}{0.8}

\begin{tabular}{l cc}
\toprule
\textbf{Data Synthesis Method} & \multicolumn{2}{c}{\textbf{STR (Hindi)}} \\
\cmidrule(l){2-3}
 & \textbf{Gemma3-1b} & \textbf{Gemma3-27b} \\
\midrule
Base                                      & 79.1 & 81.0 \\
\makecell[l]{LoRA TA trained on English data}           & 80.2 & 82.5 \\
\makecell[l]{Crosslingual TA (Hindi)\\via Adamergex}     & \textbf{80.5} & \textbf{82.7} \\
\bottomrule
\end{tabular}
\caption{Spearman correlations for STR performance in Hindi, using different data synthesis methods with 50,000 synthesized samples.}
\label{tab:str_hindi}
\end{table}

\begin{table}[H]
\centering
\scriptsize
\setlength{\tabcolsep}{5pt}
\renewcommand{\arraystretch}{0.7}

\begin{tabular}{l l c}
\toprule
\textbf{TA Adapter Specs} & \textbf{LA Adapter Specs} & \textbf{\makecell[l]{STR\\(Hindi)}} \\
\midrule
\multirow{4}{*}{\makecell[l]{Trained a combined model\\with both negative and \\positive prompt examples}}
  & Aya annotated dataset                    & 81.1 \\
  & Combined sentence corpora                & 82.0 \\
  & Individual sentences                     & 81.8 \\
  & Aya annotated + collections dataset      & 82.4 \\
\midrule
\multirow{4}{*}{\makecell[l]{Trained two individual\\models with negative and \\positive prompt examples}}
  & Aya annotated dataset                    & \textbf{82.5} \\
  & Combined sentence corpora                & \textbf{83.0} \\
  & Individual sentences                     & 81.8 \\
  & Aya annotated + collections dataset      & \textbf{82.7} \\
\bottomrule
\end{tabular}%
\caption{Spearman correlations for STR performance in Hindi, using different language adapter (LA) and task adapter (TA) specifications with 50,000 synthesized data samples. Aya annotated dataset contains the human annotated samples \citep{singh2024aya}, collections dataset denotes the Aya collections of machine translated samples. Combined sentence corpora and individual sentences experiments source the multilingual corpora from \cite{banyan}}
\label{tab:str_hindi_groups}
\end{table}

\begin{table}[H]
\centering
\scriptsize
\setlength{\tabcolsep}{6pt}
\renewcommand{\arraystretch}{0.8}

\begin{tabular}{l c c c}
\toprule
\textbf{LA Adapter Specs} & \multicolumn{3}{c}{\textbf{STR}} \\
\cmidrule(l){2-4}
 & \textbf{Hindi} & \textbf{Telugu} & \textbf{Afrikaans} \\
\midrule
Aya annotated dataset & 82.2 & 81.0 & -- \\
Combined sentence corpora & 82.7 & \textbf{81.4} & 77.4 \\
Aya annotated + collections dataset & \textbf{83.0} & 81.3 & \textbf{77.5} \\
\bottomrule
\end{tabular}%

\caption{Spearman correlations for STR performance across Hindi, Telugu, and Afrikaans for different language adapter (LA) and task adapter (TA) specifications synthesising the full dataset.  Data for Afrikaans in the Aya annotated category was unavailable.}
\label{tab:str_grouped}
\end{table}

\begin{table}[H]
\centering
\scriptsize
\setlength{\tabcolsep}{6pt}
\renewcommand{\arraystretch}{0.8}
\begin{tabular}{l c c c}
\toprule
& \multicolumn{3}{c}{\textbf{STR}} \\
\cmidrule(l){2-4}
& \textbf{Hindi} & \textbf{Telugu} & \textbf{Afrikaans} \\
\midrule
Rank 4  & 82.1 & 81.7 & 77.0 \\
Rank 8  & \textbf{83.0} & 81.3 & \textbf{77.5} \\
Rank 16 & 82.4 & \textbf{81.9} & 77.4 \\
\bottomrule
\end{tabular}%
\caption{Spearman correlations for STR performance across Hindi, Telugu, and Afrikaans for different ranks synthesising the full dataset.}
\label{tab:str_ranks}
\end{table}

\clearpage
\onecolumn
\subsection{XL-LoRA Triplet Data Synthesis Ablations}
\label{subsec:cllora_ablation}

\begin{table}[H]
\centering
\scriptsize
\setlength{\tabcolsep}{2pt}
\renewcommand{\arraystretch}{0.8}
\begin{tabular}{l ccccccc ccccccc c}
\toprule
& \multicolumn{7}{c}{\textbf{XLM-R}} & \multicolumn{7}{c}{\textbf{mmBERT}} & \textbf{Score} \\
\cmidrule(lr){2-8}\cmidrule(lr){9-15}

\textbf{Method}
& \textbf{Afr} & \textbf{Hin} & \textbf{Mar} & \textbf{Tel}
& \textbf{Ind} & \textbf{Hau} & \textbf{Kor}
& \textbf{Afr} & \textbf{Hin} & \textbf{Mar} & \textbf{Tel}
& \textbf{Ind} & \textbf{Hau} & \textbf{Kor}
& \textbf{} \\
\midrule

MT
& 80.0 & 78.2 & 83.6 & \textbf{84.7} & 48.3 & \textbf{61.0} & \textbf{73.9}
& 80.5 & \textbf{79.8} & 83.9 & 83.1 & 50.7 & \textbf{59.4} & \textbf{74.1}
& 72.9 \\

HT + MT
& 80.2 & \textbf{78.4} & 83.5 & 84.1 & \textbf{49.3} & 60.7 & 72.9
& \textbf{80.7} & \textbf{79.8} & 84.5 & \textbf{84.6} & \textbf{51.7} & 59.1 & 72.7
& \textbf{73.0} \\

HT + Syn
& \textbf{81.2} & 77.8 & \textbf{84.0} & 84.0 & 47.8 & 58.0 & 72.9
& 80.6 & 79.5 & \textbf{85.0} & 83.7 & 48.9 & 57.7 & 72.9
& 72.4 \\

\bottomrule
\end{tabular}
\caption{STR results on synthesised data methods using XL-LoRA models trained on machine-translated (MT), human-translated plus machine-translated (HT+MT) and human-translated plus synthesised (HT+Syn) data.}
\label{tab:str_clloramix_ablation}
\end{table}

\begin{table}[H]
\centering
\scriptsize
\setlength{\tabcolsep}{2pt}
\renewcommand{\arraystretch}{0.85}
\begin{tabular}{l cccc cccc c @{\hspace{6pt}\vrule\hspace{6pt}} cccc cccc c}
\toprule
& \multicolumn{9}{c}{\textbf{nDCG@10}} & \multicolumn{9}{c}{\textbf{Recall@10}} \\
\cmidrule(lr){2-10}\cmidrule(lr){11-19}

& \multicolumn{4}{c}{\textbf{XLM-R}} & \multicolumn{4}{c}{\textbf{mmBERT}} & \textbf{Score}
& \multicolumn{4}{c}{\textbf{XLM-R}} & \multicolumn{4}{c}{\textbf{mmBERT}} & \textbf{Score} \\
\cmidrule(lr){2-5}\cmidrule(lr){6-9}\cmidrule(lr){11-14}\cmidrule(lr){15-18}

\textbf{Method}
& \textbf{Hin} & \textbf{Tel} & \textbf{Ind} & \textbf{Kor}
& \textbf{Hin} & \textbf{Tel} & \textbf{Ind} & \textbf{Kor} & \textbf{}
& \textbf{Hin} & \textbf{Tel} & \textbf{Ind} & \textbf{Kor}
& \textbf{Hin} & \textbf{Tel} & \textbf{Ind} & \textbf{Kor} & \textbf{} \\
\midrule

MT
& 17.2 & 13.4 & 14.3 & 25.4
& 16.2 &  8.1 & 18.7 & 26.3 & 17.5
& 26.9 & 22.4 & 21.4 & 35.9
& 23.4 & 14.3 & 25.8 & 35.8 & 25.7 \\

HT + MT
& 18.2 & 16.6 & 17.1 & 25.2
& 18.3 & \textbf{11.9} & 19.0 & \textbf{28.9} & 19.4
& \textbf{29.3} & 26.4 & 24.2 & 34.6
& 27.1 & \textbf{20.1} & 25.7 & \textbf{37.7} & 28.1 \\

HT + Syn
& \textbf{18.8} & \textbf{16.7} & \textbf{18.1} & \textbf{25.9}
& \textbf{20.8} & 11.6 & \textbf{22.5} & 26.9 & \textbf{20.2}
& 28.4 & \textbf{28.0} & \textbf{25.2} & \textbf{37.1}
& \textbf{29.3} & 19.4 & \textbf{29.8} & 36.2 & \textbf{29.2} \\

\bottomrule
\end{tabular}
\caption{MIRACL Retrieval Hard Negative results on synthesised data methods (XL-LoRA; MT vs. HT+MT vs. HT+Syn)}
\label{tab:MIRACLRetrievalHardNegatives_clloramix_ablation}
\end{table}

\begin{table}[H]
\centering
\scriptsize
\setlength{\tabcolsep}{2pt}
\renewcommand{\arraystretch}{0.85}
\begin{tabular}{l ccc ccc c @{\hspace{6pt}\vrule\hspace{6pt}} ccc ccc c}
\toprule
& \multicolumn{7}{c}{\textbf{nDCG@10}} & \multicolumn{7}{c}{\textbf{Recall@10}} \\
\cmidrule(lr){2-8}\cmidrule(lr){9-15}

& \multicolumn{3}{c}{\textbf{XLM-R}} & \multicolumn{3}{c}{\textbf{mmBERT}} & \textbf{Score}
& \multicolumn{3}{c}{\textbf{XLM-R}} & \multicolumn{3}{c}{\textbf{mmBERT}} & \textbf{Score} \\
\cmidrule(lr){2-4}\cmidrule(lr){5-7}\cmidrule(lr){9-11}\cmidrule(lr){12-14}

\textbf{Method}
& \textbf{Hin} & \textbf{Mar} & \textbf{Tel}
& \textbf{Hin} & \textbf{Mar} & \textbf{Tel} & \textbf{}
& \textbf{Hin} & \textbf{Mar} & \textbf{Tel}
& \textbf{Hin} & \textbf{Mar} & \textbf{Tel} & \textbf{} \\
\midrule

MT
& 49.9 & 56.8 & 54.2
& 46.6 & 55.7 & 49.5 & 52.1
& 69.4 & 75.0 & 74.6
& 65.2 & 74.3 & 67.8 & 71.1 \\

HT + MT
& 50.2 & 57.5 & 54.9
& 48.2 & 56.2 & 51.1 & 53.0
& 68.7 & 75.3 & 74.7
& 67.2 & 75.0 & 68.9 & 71.6 \\

HT + Syn
& \textbf{51.2} & \textbf{57.6} & \textbf{55.1}
& \textbf{48.8} & \textbf{58.6} & \textbf{51.6} & \textbf{53.8}
& \textbf{70.6} & \textbf{75.8} & \textbf{75.4}
& \textbf{68.2} & \textbf{76.8} & \textbf{69.7} & \textbf{72.8} \\

\bottomrule
\end{tabular}%
\caption{Indic QA Retrieval results on synthesised data (XL-LoRA; MT vs. HT+MT vs. HT+Syn).}
\label{tab:indicqaretrieval_clloramix_ablation}
\end{table}

\begin{table}[H]
\centering
\scriptsize
\setlength{\tabcolsep}{2pt}
\renewcommand{\arraystretch}{0.85}
\begin{tabular}{l ccccccc ccccccc c}
\toprule
\textbf{nDCG@10}
& \multicolumn{7}{c}{\textbf{XLM-R}}
& \multicolumn{7}{c}{\textbf{mmBERT}}
& \textbf{Score} \\
\cmidrule(lr){2-8}\cmidrule(lr){9-15}

\textbf{Method}
& \textbf{Afr} & \textbf{Hin} & \textbf{Mar} & \textbf{Tel}
& \textbf{Ind} & \textbf{Hau} & \textbf{Kor}
& \textbf{Afr} & \textbf{Hin} & \textbf{Mar} & \textbf{Tel}
& \textbf{Ind} & \textbf{Hau} & \textbf{Kor}
& \textbf{} \\
\midrule

MT
& 76.8 & 68.2 & 69.6 & 63.6 & 77.7 & 65.4 & 75.4
& 78.7 & 69.4 & 70.5 & 65.7 & 82.4 & \textbf{66.3} & 79.1
& 72.1 \\

HT + MT
& 78.0 & \textbf{70.1} & \textbf{71.8} & 64.4 & 78.9 & 64.6 & 76.3
& 81.1 & \textbf{70.8} & 72.6 & \textbf{65.6} & 82.0 & 62.3 & 80.4
& 72.8 \\

HT + Syn
& \textbf{79.2} & \textbf{70.1} & 71.6 & \textbf{64.6}
& \textbf{79.1} & \textbf{65.8} & \textbf{78.4}
& \textbf{81.9} & \textbf{70.8} & \textbf{73.9} & 65.5
& \textbf{83.7} & 58.9 & \textbf{81.0}
& \textbf{73.2} \\

\bottomrule
\end{tabular}
\caption{Belebele Retrieval nDCG@10 results on synthesised data (XL-LoRA; MT vs. HT+MT vs. HT+Syn)}
\label{tab:belebele_ndcg_clloramix_ablation}
\end{table}

\begin{table}[H]
\centering
\scriptsize
\setlength{\tabcolsep}{2pt}
\renewcommand{\arraystretch}{0.85}
\begin{tabular}{l ccccccc ccccccc c}
\toprule
\textbf{Recall@10}
& \multicolumn{7}{c}{\textbf{XLM-R}}
& \multicolumn{7}{c}{\textbf{mmBERT}}
& \textbf{Score} \\
\cmidrule(lr){2-8}\cmidrule(lr){9-15}

\textbf{Method}
& \textbf{Afr} & \textbf{Hin} & \textbf{Mar} & \textbf{Tel}
& \textbf{Ind} & \textbf{Hau} & \textbf{Kor}
& \textbf{Afr} & \textbf{Hin} & \textbf{Mar} & \textbf{Tel}
& \textbf{Ind} & \textbf{Hau} & \textbf{Kor}
& \textbf{} \\
\midrule

MT
& 88.9 & 84.1 & 84.6 & 80.1 & 90.1 & 79.0 & 89.4
& 90.4 & 86.1 & 85.3 & 80.6 & 93.4 & \textbf{79.8} & 91.8
& 86.0 \\

HT + MT
& 90.1 & 84.7 & 86.8 & \textbf{81.1} & 91.0 & 79.1 & 89.7
& 92.6 & 85.2 & 86.4 & 80.2 & 93.0 & 79.2 & 92.1
& 86.5 \\

HT + Syn
& \textbf{91.1} & \textbf{84.8} & \textbf{87.0} & 80.2
& \textbf{91.9} & \textbf{80.1} & \textbf{91.6}
& \textbf{92.7} & \textbf{86.3} & \textbf{87.8} & \textbf{81.2}
& \textbf{94.0} & 74.2 & \textbf{93.3}
& \textbf{86.9} \\

\bottomrule
\end{tabular}
\caption{Belebele Retrieval Recall@10 results on synthesised data (XL-LoRA; MT vs. HT+MT vs. HT+Syn)}
\label{tab:belebele_recall_clloramix_ablation}
\end{table}

\clearpage
\subsection{Retrieval evaluation}
\label{subsec:retrieval_eval}

\begin{table}[H]
\centering
\scriptsize
\setlength{\tabcolsep}{3pt}
\renewcommand{\arraystretch}{1.05}
\begin{tabular}{l cccc cccc c @{\hspace{10pt}\vrule\hspace{10pt}} cccc cccc c}
\toprule
& \multicolumn{9}{c}{\textbf{nDCG@10}} & \multicolumn{9}{c}{\textbf{Recall@10}} \\
\cmidrule(lr){2-10}\cmidrule(lr){11-19}

& \multicolumn{4}{c}{\textbf{XLM-R}} & \multicolumn{4}{c}{\textbf{mmBERT}} & \textbf{Score}
& \multicolumn{4}{c}{\textbf{XLM-R}} & \multicolumn{4}{c}{\textbf{mmBERT}} & \textbf{Score} \\
\cmidrule(lr){2-5}\cmidrule(lr){6-9}\cmidrule(lr){11-14}\cmidrule(lr){15-18}

\textbf{Method}
& \textbf{Hin} & \textbf{Tel} & \textbf{Ind} & \textbf{Kor}
& \textbf{Hin} & \textbf{Tel} & \textbf{Ind} & \textbf{Kor} & \textbf{}
& \textbf{Hin} & \textbf{Tel} & \textbf{Ind} & \textbf{Kor}
& \textbf{Hin} & \textbf{Tel} & \textbf{Ind} & \textbf{Kor} & \textbf{} \\
\midrule

Base Encoder
& 1.4 & 0.6 & 0.3 & 4.9
& 0.3 & 0.0 & 0.0 & 0.7 & 1.0
& 1.7 & 0.7 & 0.6 & 6.6
& 0.4 & 0.0 & 0.1 & 0.7 & 1.4 \\

Unsupervised
& 16.1 & 11.0 & 8.6 & 22.7
& 2.2 & 0.7 & 3.3 & 7.7 & 9.0
& 23.5 & 17.6 & 12.1 & 29.1
& 3.8 & 1.0 & 5.2 & 12.1 & 13.1 \\

Cross Lingual
& \textbf{19.8} & 13.3 & 16.5 & \textbf{27.4}
& \textbf{22.0} & 8.2 & 20.0 & 28.3 & 19.4
& \textbf{29.8} & 22.0 & 23.9 & 36.3
& \textbf{30.6} & 13.6 & 27.7 & 37.5 & 27.7 \\

Synth - Prompting
& 16.7 & 16.3 & 15.1 & 21.6
& 9.5 & 11.4 & 9.7 & 24.5 & 15.6
& 24.5 & \textbf{28.8} & 21.6 & 32.5
& 13.6 & \textbf{20.2} & 14.5 & 36.6 & 24.0 \\

Synth - Adapter Composition
& 16.5 & \textbf{17.0} & \textbf{18.8} & 25.5
& 16.6 & 11.5 & 21.7 & \textbf{29.1} & 19.6
& 26.5 & 28.4 & \textbf{26.4} & 37.4
& 25.4 & 19.8 & 29.4 & \textbf{40.5} & \textbf{29.2} \\

Synth - XL-LoRA
& 18.8 & 16.7 & 18.1 & 25.9
& 20.8 & \textbf{11.6} & \textbf{22.5} & 26.9 & \textbf{20.2}
& 28.4 & 28.0 & 25.2 & \textbf{37.1}
& 29.3 & 19.4 & \textbf{29.8} & 36.2 & \textbf{29.2} \\

\bottomrule
\end{tabular}
\caption{MIRACL Retrieval Hard Negative results across languages.}
\label{tab:miracl_fullstyle_ndcg_recall}
\end{table}

\begin{table}[H]
\centering
\scriptsize
\setlength{\tabcolsep}{3pt}
\renewcommand{\arraystretch}{1.05}
\begin{tabular}{l ccc ccc c @{\hspace{10pt}\vrule\hspace{10pt}} ccc ccc c}
\toprule
& \multicolumn{7}{c}{\textbf{nDCG@10}} & \multicolumn{7}{c}{\textbf{Recall@10}} \\
\cmidrule(lr){2-8}\cmidrule(lr){9-15}

& \multicolumn{3}{c}{\textbf{XLM-R}} & \multicolumn{3}{c}{\textbf{mmBERT}} & \textbf{Score}
& \multicolumn{3}{c}{\textbf{XLM-R}} & \multicolumn{3}{c}{\textbf{mmBERT}} & \textbf{Score} \\
\cmidrule(lr){2-4}\cmidrule(lr){5-7}\cmidrule(lr){9-11}\cmidrule(lr){12-14}

\textbf{Method}
& \textbf{Hin} & \textbf{Mar} & \textbf{Tel}
& \textbf{Hin} & \textbf{Mar} & \textbf{Tel} & \textbf{}
& \textbf{Hin} & \textbf{Mar} & \textbf{Tel}
& \textbf{Hin} & \textbf{Mar} & \textbf{Tel} & \textbf{} \\
\midrule

Base Encoder
& 5.9 & 21.0 & 21.7
& 1.9 & 2.0 & 39.6 & 15.4
& 12.2 & 31.1 & 43.1
& 4.3 & 4.6 & 57.2 & 25.4 \\

Unsupervised
& 43.2 & 44.6 & 48.2
& 17.1 & 23.1 & 13.9 & 31.7
& 61.9 & 62.8 & 67.2
& 31.5 & 38.7 & 24.2 & 47.7 \\

Cross Lingual
& 50.7 & 55.0 & 50.8
& 46.2 & 50.1 & 39.6 & 48.7
& 70.0 & 73.4 & 69.6
& 64.9 & 68.3 & 57.2 & 67.2 \\

Synth - Prompting
& 52.3 & 52.3 & 51.8
& 44.4 & 47.8 & 42.5 & 48.5
& \textbf{72.0} & 71.7 & 72.1
& 62.9 & 67.0 & 59.9 & 67.6 \\

Synth - Adapter Composition
& \textbf{52.5} & 57.5 & \textbf{57.7}
& 48.5 & 52.7 & 47.0 & 52.7
& 71.5 & \textbf{75.8} & \textbf{77.5}
& 66.2 & 72.3 & 65.1 & 71.4 \\

Synth - XL-LoRA
& 51.2 & \textbf{57.6} & 55.1
& \textbf{48.8} & \textbf{58.6} & \textbf{51.6} & \textbf{53.8}
& 70.6 & \textbf{75.8} & 75.4
& \textbf{68.2} & \textbf{76.8} & \textbf{69.7} & \textbf{72.8} \\

\bottomrule
\end{tabular}
\caption{Indic QA Retrieval results across languages}
\label{tab:indicqa_style_ndcg_recall}
\end{table}

\begin{table}[H]
\centering
\scriptsize
\setlength{\tabcolsep}{3pt}
\renewcommand{\arraystretch}{1.05}
\begin{tabular}{l ccccccc ccccccc c}
\toprule
\textbf{nDCG@10}
& \multicolumn{7}{c}{\textbf{XLM-R}}
& \multicolumn{7}{c}{\textbf{mmBERT}}
& \textbf{Score} \\
\cmidrule(lr){2-8}\cmidrule(lr){9-15}

\textbf{Method}
& \textbf{Afr} & \textbf{Hin} & \textbf{Mar} & \textbf{Tel}
& \textbf{Ind} & \textbf{Hau} & \textbf{Kor}
& \textbf{Afr} & \textbf{Hin} & \textbf{Mar} & \textbf{Tel}
& \textbf{Ind} & \textbf{Hau} & \textbf{Kor}
& \textbf{} \\
\midrule

Base Encoder
& 14.7 & 11.5 & 13.8 & 14.8 & 27.9 & 5.9 & 23.0
& 1.1 & 0.8 & 0.9 & 1.0 & 1.1 & 1.2 & 1.2
& 8.5 \\

Unsupervised
& 7.4 & 8.0 & 5.7 & 49.7 & 66.5 & 18.6 & 62.4
& 36.1 & 20.1 & 16.8 & 12.5 & 54.4 & 5.0 & 32.1
& 28.2 \\

Cross Lingual
& 74.9 & 65.6 & 66.8 & 60.2 & \textbf{79.1} & 41.9 & 74.3
& 78.5 & 64.7 & 63.0 & 48.0 & 82.9 & 30.0 & 77.4
& 64.8 \\

Synth - Prompting
& 70.5 & 58.5 & 61.0 & 56.6 & 71.0 & 48.3 & 72.7
& 65.0 & 53.6 & 53.5 & 48.7 & 73.2 & 28.0 & 70.9
& 59.4 \\

Synth - Adapter Composition
& 77.7 & 64.0 & 68.6 & 61.3 & 76.9 & 56.7 & 73.1
& 76.6 & 58.8 & 61.0 & 57.6 & 79.9 & 43.6 & 72.7
& 66.3 \\

Synth - XL-LoRA
& \textbf{79.2} & \textbf{70.1} & \textbf{71.6} & \textbf{64.6}
& \textbf{79.1} & \textbf{65.8} & \textbf{78.4}
& \textbf{81.9} & \textbf{70.8} & \textbf{73.9} & \textbf{65.5}
& \textbf{83.7} & \textbf{58.9} & \textbf{81.0}
& \textbf{73.2} \\

\bottomrule
\end{tabular}
\caption{Belebele Retrieval nDCG@10 results across languages}
\label{tab:belebele_ndcg_full_combined}
\end{table}

\begin{table}[H]
\centering
\scriptsize
\setlength{\tabcolsep}{3pt}
\renewcommand{\arraystretch}{1.05}
\begin{tabular}{l ccccccc ccccccc c}
\toprule
\textbf{Recall@10}
& \multicolumn{7}{c}{\textbf{XLM-R}}
& \multicolumn{7}{c}{\textbf{mmBERT}}
& \textbf{Score} \\
\cmidrule(lr){2-8}\cmidrule(lr){9-15}

\textbf{Method}
& \textbf{Afr} & \textbf{Hin} & \textbf{Mar} & \textbf{Tel}
& \textbf{Ind} & \textbf{Hau} & \textbf{Kor}
& \textbf{Afr} & \textbf{Hin} & \textbf{Mar} & \textbf{Tel}
& \textbf{Ind} & \textbf{Hau} & \textbf{Kor}
& \textbf{} \\
\midrule

Base Encoder
& 20.8 & 16.8 & 20.4 & 20.9 & 39.8 & 10.6 & 31.9
& 2.2 & 1.7 & 2.0 & 2.1 & 2.1 & 2.4 & 2.4
& 12.6 \\

Unsupervised
& 12.8 & 13.7 & 11.2 & 64.1 & 81.2 & 30.0 & 78.3
& 52.2 & 32.0 & 27.0 & 22.0 & 69.2 & 8.9 & 47.0
& 39.3 \\

Cross Lingual
& 87.4 & 80.0 & 83.1 & 77.1 & 90.3 & 54.2 & 86.9
& 90.6 & 78.7 & 78.1 & 62.8 & 92.7 & 37.3 & 90.7
& 77.9 \\

Synth - Prompting
& 84.1 & 73.4 & 77.3 & 71.7 & 84.6 & 63.8 & 85.9
& 81.1 & 69.6 & 70.8 & 64.9 & 86.6 & 40.9 & 84.4
& 74.2 \\

Synth - Adapter Composition
& 89.1 & 78.2 & 83.8 & 75.7 & 89.3 & 72.4 & 85.9
& 90.4 & 75.6 & 76.6 & 72.4 & 90.7 & 58.8 & 85.1
& 80.3 \\

Synth - XL-LoRA
& \textbf{91.1} & \textbf{84.8} & \textbf{87.0} & \textbf{80.2}
& \textbf{91.9} & \textbf{80.1} & \textbf{91.6}
& \textbf{92.7} & \textbf{86.3} & \textbf{87.8} & \textbf{81.2}
& \textbf{94.0} & \textbf{74.2} & \textbf{93.3}
& \textbf{86.9} \\

\bottomrule
\end{tabular}
\caption{Belebele Retrieval Recall@10 results across languages}
\label{tab:belebele_recall_full_combined}
\end{table}

\subsection{Zero-shot Prompting Strategies for Task Adapters}
\label{subsec:prompting_adamergex}

\begin{table}[H]
\centering
\scriptsize
\setlength{\tabcolsep}{6pt}
\renewcommand{\arraystretch}{1.2}
\begin{tabular}{p{0.2\textwidth} p{0.4\textwidth}}
\toprule
\textbf{Prompting Strategy} & \textbf{Zero-Shot Prompt} \\
\midrule

\citet{huang-etal-2023-languages} Prompting -- XNLI Classification
&
I want you to act as a natural language inference expert for \{language\}. 
Premise: \{premise\}. Hypothesis: \{hypothesis\}. 
You should retell the premise and hypothesis in English. 
You should judge whether the hypothesis is true (entailment), false (contradiction), 
or undetermined (neutral) given the premise. 
The relationship can be chosen from entailment, contradiction and neutral. 
You should step-by-step answer the request. 
You should tell me the relationship in this format `Relationship:'.
\\

Modified Prompting -- XNLI Classification (This Work)
&
You are an expert at natural language inference. Given a premise and hypothesis 
(in \{language\}), you should return an integer classification. 
The options are as follows: 0 for entailment, 1 for neutral, and 2 for contradiction. 
Return only the integer without any preamble or explanation. 
Premise: \{premise\}. Hypothesis: \{hypothesis\}.
\\

Prompting -- Positive Pair Generation (This Work)
&
You are an expert at natural language inference. Given a premise, 
you should return an entailment sentence example (in \{language\}) to the premise. 
Return only the entailment sentence example without any preamble or explanation. 
Premise: \{premise\}.
\\

Prompting -- Negative Pair Generation (This Work)
&
You are an expert at natural language inference. Given a premise, 
you should return a contradiction sentence example (in \{language\}) to the premise. 
Return only the contradiction sentence example without any preamble or explanation. 
Premise: \{premise\}.
\\

\bottomrule
\end{tabular}
\caption{Zero-shot prompting strategies used in this study.}
\label{tab:zeroshot_prompting}
\end{table}

\subsection{Qualitative Analysis - Turkish data}
\label{subsec:qualitative_turkish}

\begin{figure*}[h]
\centering
\includegraphics[width=1.1\textwidth]{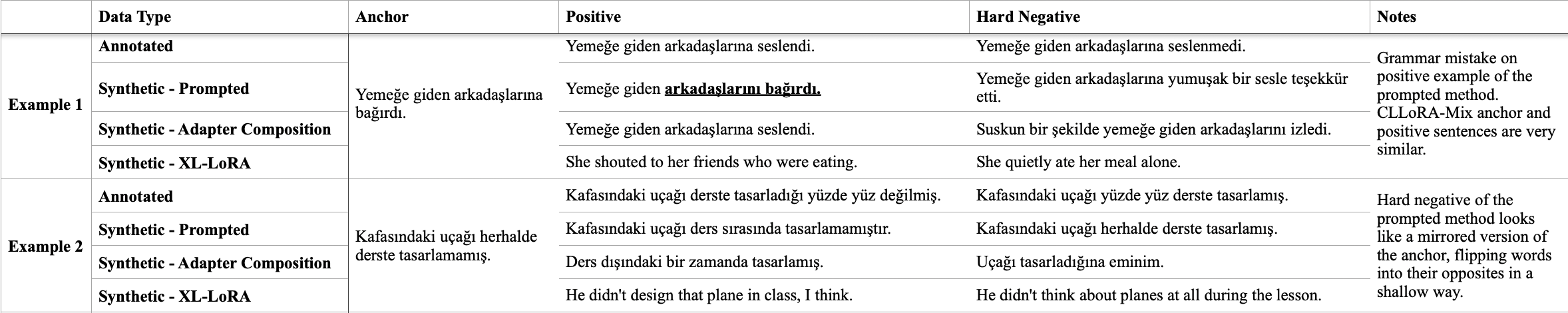}
\caption{Qualitative review comparing the Turkish semi annotated data \citep{halat-atlamaz-2024-implicatr}, synthetic data via Prompted, Adapter Composition and XL-LoRA methods.  We constructed the annotated triplet dataset from the ImplicaTR dataset \citep{halat-atlamaz-2024-implicatr} by treating entailment pairs as positive examples and contradiction pairs as hard negatives.}
\label{fig:turkish_triplet_examples}
\end{figure*}

\subsection{Alignment and Uniformity}
\label{subsec:alig_unif_tel}

\begin{figure}[H]
    \centering
    \includegraphics[width=0.3\linewidth]{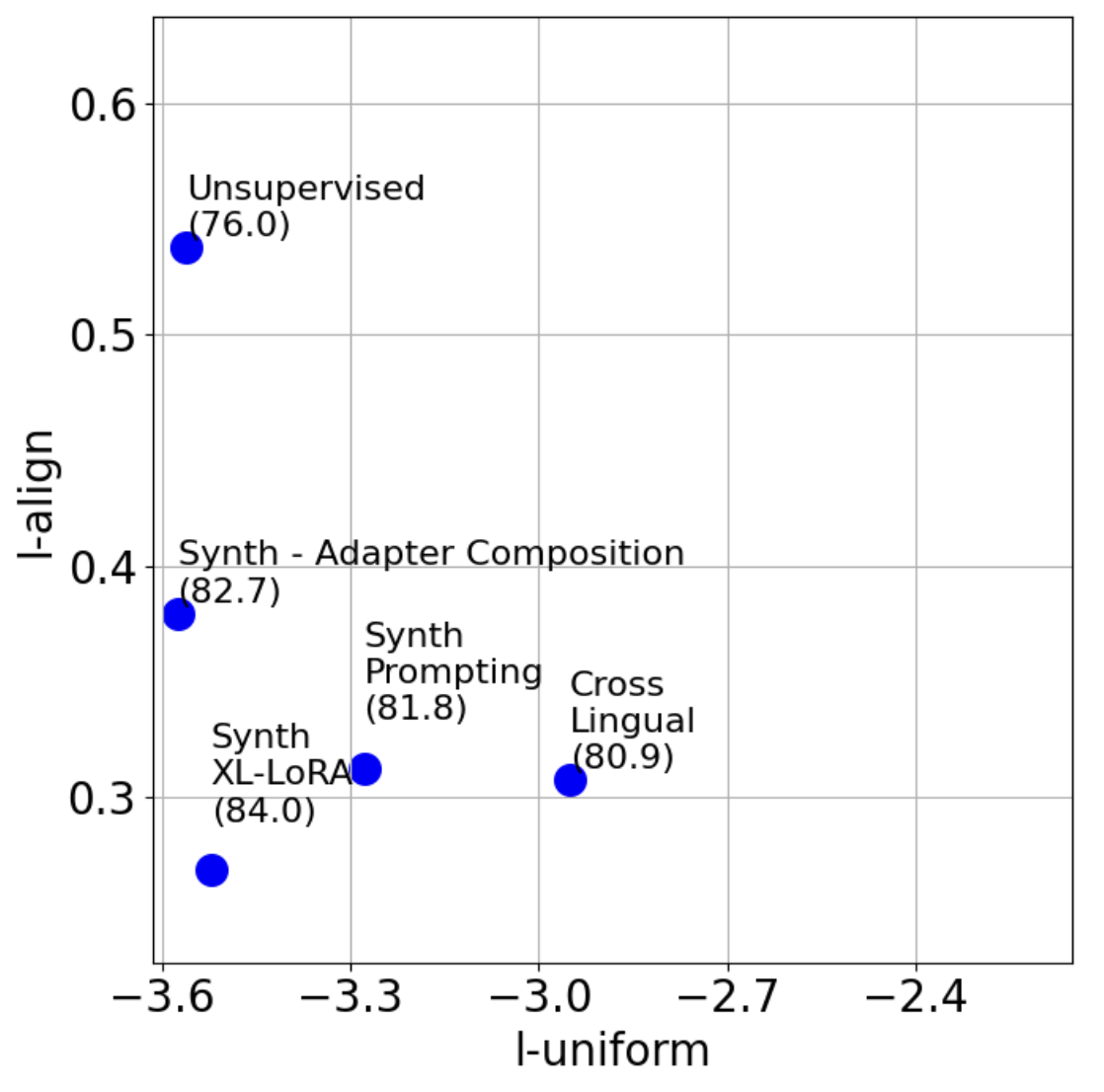}
    \caption{Alignment and Uniformity analysis of XLM-R based fine tuned sentence embedding models, based on Telugu STR language embeddings.}
    \label{fig:lalign_lunif_telugu}
\end{figure}

\end{document}